\documentclass[journal]{IEEEtran}

\usepackage{amsfonts}
\usepackage{color}
\usepackage{times}
\usepackage{epsfig}
\usepackage{graphicx}
\usepackage{amsmath}
\usepackage{amssymb}
\usepackage{multirow}
\usepackage{cite}
\usepackage{epstopdf}
\usepackage{subfigure}
\usepackage{url}

\begin{document}

\title{Context-Aware Text-Based Binary Image \\Stylization and Synthesis}

\author{Shuai~Yang,
        Jiaying~Liu,~\IEEEmembership{Senior Member,~IEEE,}
        Wenhan Yang,~\IEEEmembership{Student Member,~IEEE,}
        and~Zongming~Guo,~\IEEEmembership{Member,~IEEE,}
\thanks{
This work was supported by National Natural Science Foundation of China under contract No. 61772043.
\textit{(Corresponding author: Jiaying Liu.)}}
\thanks{The authors are with Institute of Computer Science and Technology, Peking University, Beijing 100871, China, (e-mail: \textnormal{williamyang@pku.edu.cn, liujiaying@pku.edu.cn, yangwenhan@pku.edu.cn, guozongming@pku.edu.cn).}}
\thanks{Project page: \protect\url{http://www.icst.pku.edu.cn/struct/Projects/UTS.html}}
}
\maketitle

\begin{abstract}
In this work, we present a new framework for the stylization of text-based binary images.
First, our method stylizes the stroke-based geometric shape like text, symbols and icons in the target binary image based on an input style image. Second, the composition of the stylized geometric shape and a background image is explored.
To accomplish the task, we propose legibility-preserving structure and texture transfer algorithms, which progressively narrow the visual differences between the binary image and the style image. The stylization is then followed by a context-aware layout design algorithm, where cues for both seamlessness and aesthetics are employed to determine the optimal layout of the shape in the background. Given the layout, the binary image is seamlessly embedded into the background by texture synthesis under a context-aware boundary constraint.
According to the contents of binary images, our method can be applied to many fields. We show that the proposed method is capable of addressing the unsupervised text stylization problem and is superior to state-of-the-art style transfer methods in automatic artistic typography creation. Besides, extensive experiments on various tasks, such as visual-textual presentation synthesis, icon/symbol rendering and structure-guided image inpainting, demonstrate the effectiveness of the proposed method.
\end{abstract}


\IEEEpeerreviewmaketitle

\section{Introduction}

\IEEEPARstart{S}{TYLE} transfer is the task of migrating a style from an image to another to synthesize a new artistic image. It is of special interest in visual design, and has applications such as painting synthesis and photography post-processing. However, creating an image in a particular style manually requires great skills that are beyond the capabilities of average users. Therefore, automatic style transfer has become a trending topic both in academic literature and industrial applications.

\begin{figure}
\centering
    \subfigure[Unsupervised text-based binary image stylization]{\label{fig:super}
    \includegraphics[width=0.98\linewidth]{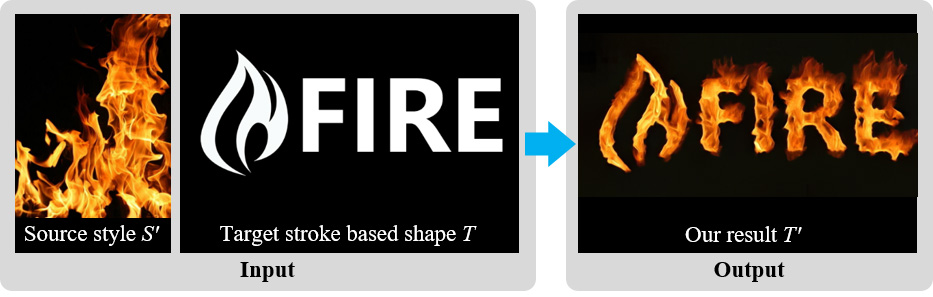}}\hfill
    \subfigure[Supervised text stylization~\cite{Yang2017Awesome}]{\label{fig:unsuper}
    \includegraphics[width=0.98\linewidth]{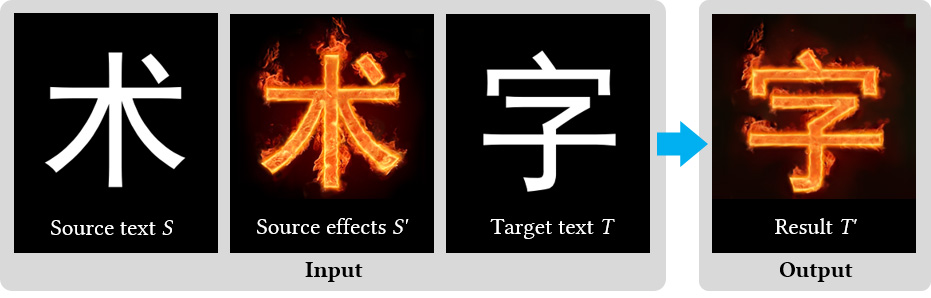}}
  \caption{Supervised text stylization requires registered raw text $S$ and text effects $S'$ as input. We instead handle a more challenging unsupervised text-based image stylization problem with an arbitrary source image $S'$. (a) Our unsupervised stylization result. (b) Supervised stylization result by~\cite{Yang2017Awesome}.}\label{fig:supervsunsuper}
\end{figure}

Text and other stroke-based design elements such as symbols, icons and labels highly summarize the abstract imagery of human visual perceptions and are ubiquitous in our daily life. The stylization of text-based binary images as in Fig.~\ref{fig:supervsunsuper}(a) is of great research value but also poses a challenge of narrowing the great visual discrepancy between the binary flat shapes and the colorful style image.

Style transfer has been investigated for years, where many successful methods are proposed, such as the non-parametric method Image Quilting~\cite{Efros2001Image} and the parametric method Neural Style~\cite{gatys2016image}. Non-parametric methods take samples from the style image and place the samples based on pixel intensity~\cite{Efros2001Image,Frigo2016Split,Elad2016Style} or deep features~\cite{Chen2017Visual} of the target image to synthesize a new image. Parametric methods represent the style as statistical features, and adjust the target image to satisfy these features. Recent deep learning based parametric methods~\cite{gatys2016image,Li2016Combining,Chen2017StyleBank} exploit high-level deep features, and thereby have the superior capability of semantic style transfer. However, none of the aforementioned methods are specific to the stylization of text-based binary images. In fact, for non-parametric methods, it is hard to use pixel intensity or deep features to establish a direct mapping between a binary image and a style image, due to their great modality discrepancy. On the other hand, text-based binary images lack high-level semantic information, which limits the performance of the parametric methods.

As the most related method to our problem, a text effects transfer algorithm~\cite{Yang2017Awesome} is recently proposed to stylize the binary text image. In that work, the authors analyzed the high correlation between texture patterns and their spatial distribution in text effects images, and modeled it as a distribution prior, which has been proven to be highly effective at text stylization. But this method strictly requires the source style to be a well-structured typography image. Moreover, it follows the idea of Image Analogies~\cite{Hertzmann2001Image} to stylize the image in a supervised manner. For supervised style transfer, in addition to the source style image, its non-stylized counterpart is also required to learn the transformation between them, as shown in Fig.~\ref{fig:supervsunsuper}(b). Unfortunately, such a pair of inputs is not readily available in practice, which greatly limits its application scope.

In this work, we handle a more challenging unsupervised stylization problem, only with a binary text-based binary image and an arbitrary style image as in Fig.~\ref{fig:supervsunsuper}(a). 
To bridge the distinct visual discrepancies between the binary image and the style image, we extract the main structural imagery of the style image to build a preliminary mapping to the binary image. The mapping is then refined using a structure transfer algorithm, which adds shape characteristics of the source style to the binary shape. In addition to the distribution constraint~\cite{Yang2017Awesome}, a saliency constraint is proposed to jointly guide the texture transfer process for the shape legibility. These improvements allow our unsupervised style transfer to yield satisfying artistic results without the ideal input required by supervised methods.

Furthermore, we investigate the combination of stylized shapes (text, symbols, icons) and background images, which is very common in visual design. Specifically, we propose a new context-aware text-based binary image stylization and synthesis framework, where the target binary shape is seamlessly embedded in a background image with a specified style. By ``seamless'', we mean the target shape is stylized to share context consistency with the background image without abrupt image boundaries, such as decorating a blue sky with cloud-like typography. To achieve it, we leverage cues considering both seamlessness and aesthetics to determine the image layout, where the target shape is finally synthesized into the background image. When a series of different styles are available, our method can generate diverse artistic typography, symbols or icons against the background image, thereby facilitating a much wider variety of aesthetic interest expression. In summary, our major technical contributions are:
\begin{itemize}
  \item We raise a new text-based binary image stylization and synthesis problem for visual design and develop the first automatic aesthetic driven framework to solve it.
  \item We present novel structure and texture transfer algorithms to balance shape legibility with texture consistency, which we show to be effective in style transition between the binary shape and the style image.
  \item We propose a context-aware layout design method to create professional looking artwork, which determines the image layout and seamlessly synthesizes the artistic shape into the background image.
\end{itemize}

The rest of this paper is organized as follows.
In Section~\ref{sec:related_work}, we review related works in style transfer and text editing.
Section~\ref{sec:overview} defines the text-based binary image stylization problem, and gives an overview of the framework of our method.
In Section~\ref{sec:text_style_transfer} and \ref{sec:text_embedding}, the details of the proposed legibility-preserving style transfer method and context-aware layout design method are presented, respectively.
We validate our method by conducting extensive experiments and comparing with state-of-the-art style transfer algorithms in Section~\ref{sec:experiment}.
Finally, we conclude our work in Section~\ref{sec:conclusion}.

\section{Related Work}
\label{sec:related_work}

\subsection{Color Transfer}

Pioneering methods transfer colors by applying a global transformation to the target image to match the color statistics of a source image~\cite{Reinhard2001Color,hertzmann2001algorithms,Piti2007Automated}. When the target image and the source image have similar content, these methods generate satisfying results. Subsequent methods work on color transfer in a local manner to cope with the images of arbitrary scenes. They infer local color statistics in different regions by image segmentation~\cite{Tai2005Local,Tai2007Soft}, perceptual color categories~\cite{chang2006example,Chang2007Example} or user interaction~\cite{Welsh2002Transferring}. More recently, Shih~\textit{et al}. employed fine-grained patch/pixel correspondences to transfer illumination and color styles for landscape images~\cite{Shih2013Data} and headshot portraits~\cite{Shih2014Style}. Yan~\textit{et al}.~\cite{Yan2016Automatic} leveraged deep neural networks to learn effective color transforms from a large database. In this paper, we employ color transfer technology~\cite{hertzmann2001algorithms} to reduce the color difference between the style image and the background image for seamless shape embedding.

\begin{figure*}
  \centering
    \includegraphics[width=0.99\linewidth]{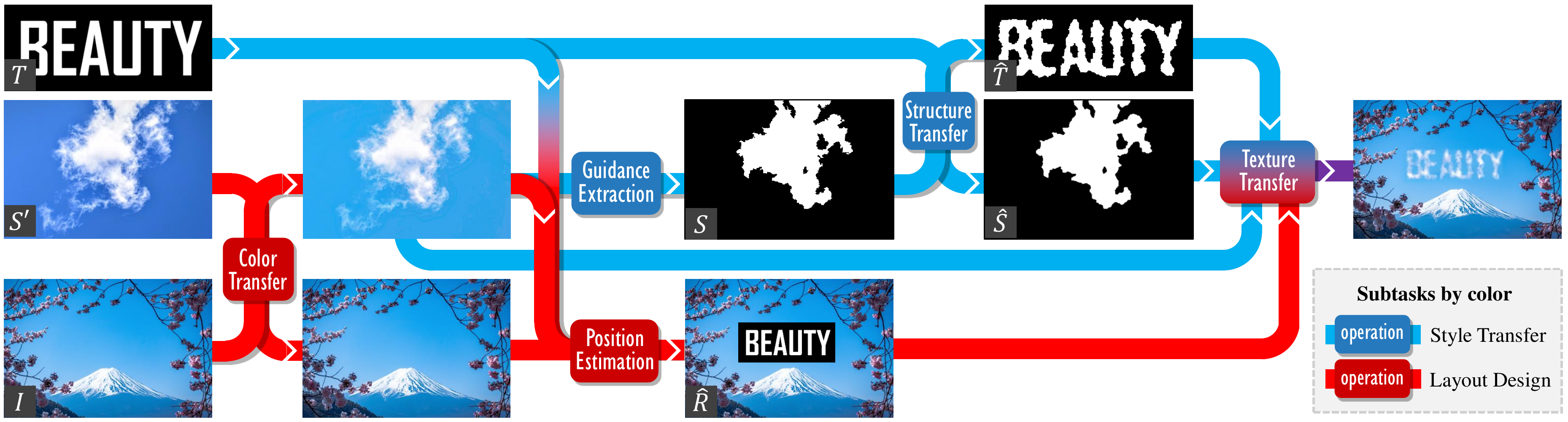}
  \caption{Overview of the proposed algorithm. Our method consists of two subtasks: style transfer for stylizing the target text-based binary image $T$ based on a style image $S'$ and layout design for synthesizing the stylized image into a background image $I$. We first extract a guidance map $S$ from $S'$ to establish a mapping with $T$. Then a structure adjustment is adopted to bridge the structural gap between $T$ and $S$. Under the guidance of $\hat{S}$ and $\hat{T}$, the stylized image is generated via texture transfer. To embed the target shape of $T$ into the background image, we develop a method to determine the image layout $\hat{R}$. The colors of $I$ and $S'$ are adjusted and the contextual information of $I$ is used to constrain the texture transfer for visual seamlessness. This flowchart gives an example of synthesizing a visual-textual presentation when $T$ is a text image. The operation modules are colored to show the subtasks they belong to, where dark blue and dark red represent style transfer and layout design, respectively. Best viewed in color. \textit{Image credits: Unsplash users JJ Ying, Tim Gouw\protect\footnotemark[1].}
  }\label{fig:overview}
\end{figure*}

\subsection{Texture Synthesis}

Texture synthesis technologies attempt to generate new textures from a given texture example. Non-parametric methods use pixel~\cite{efros1999texture} or patch~\cite{Efros2001Image} samplings in the example to synthesize new textures. For these methods, the coherence of neighboring samples is the research focus, where patch blending via image averaging~\cite{Liang2001Real}, dynamic programming~\cite{Efros2001Image}, graph cut~\cite{Kwatra2003Graphcut} and coherence function optimization~\cite{Space-time} is proposed. Meanwhile, parametric methods build mathematic models to simulate certain texture statistics of the texture example. Among this kind of methods, the most popular one is the Gram-matrix model proposed by Gatys \textit{et al}.~\cite{gatys2015texture}. Using the correlations between multi-level deep features to represent textures, this model produces natural textures of noticeably high perceptual quality. In this paper, we adapt conventional texture synthesis methods to dealing with binary text images. We apply four constrains of text shape, texture distribution, texture repetitiveness and text saliency to the texture synthesis method of Wexler~\cite{Space-time} to build our novel texture transfer model.

\subsection{Texture Transfer}

In texture transfer, textures are synthesized under the structure constraint from an additional content image. According to whether a guidance map is provided, texture transfer can be further categorized into supervised and unsupervised methods.

Supervised methods, also known as image analogies~\cite{Hertzmann2001Image}, rely on the availability of an input image and its stylized result. These methods learn a mapping between such an example pair, and stylize the target image by applying the learned mapping to it. Since first reported in~\cite{Hertzmann2001Image}, image analogies have been extended in various ways such as video analogies~\cite{Nard2015Stylizing} and fast image analogies~\cite{Barnes2015PatchTable}. The main drawback of image analogies is the strict requirement for the registered example pair. Most often, we only have a style image at hand, and need to turn to the unsupervised texture transfer methods.

Without the guidance of the example pair, unsupervised methods directly find mappings between different texture modalities. For instance, Efros and Freeman \cite{Efros2001Image} introduced a guidance map derived from image intensity to help find correspondences between two texture modalities. Zhang \textit{et al}.~\cite{Zhang2015Robust} used a sparse-based initial sketch estimation~\cite{Zhang2015Face} to construct a mapping between the source sketch texture and the target image. Frigo~\textit{et~al}.~\cite{Frigo2016Split} put forward a patch partition mechanism for an adaptive patch mapping, which balances the preservation of structures and textures. However, these methods attempt to use intensity features to establish a direct mapping between the target image and the style image, and will fail in our case where the two input images have huge visual differences. By contrast, our method proposes to extract an abstract binary imagery from the style image, which shares the same modality as the target image and serves as a bridge.

Fueled by the recent development of deep learning, there has been rapid advancement of deep-based methods that leverage high-level image features for style transfer. In pioneering Neural Style~\cite{gatys2016image}, the authors adapted Gram-matrix-based texture synthesis~\cite{gatys2015texture} to style transfer by incorporating content similarities, which enables the composition of different perceptual information. This method has inspired a new wave of research on video stylization~\cite{Chen2017Coherent}, perceptual factor control~\cite{Gatys2017Controlling} and acceleration~\cite{Johnson2016Perceptual}. In parallel, Li and Wand~\cite{Li2016Combining} introduced a framework called CNNMRF that exploits Markov Random Field (MRF) to enforce local texture transfer. Based on CNNMRF, Neural Doodle~\cite{Champandard2016Semantic} incorporates semantic maps for analogy guidance, which turns semantic maps into artwork.
The main advantage of parametric deep-based methods is their ability to establish semantic mappings. For instance, it is reported in~\cite{Li2016Combining} that the network can find accurate correspondences between real faces and sketched faces, even if their appearances differ greatly in pixel domain. However, in our problem, the plain text image provides little semantic information, making these parametric methods lose their advantages in comparison to our non-parametric method, as demonstrated in Fig.~\ref{fig:experiment1}.

\subsection{Text Stylization}

In the domain of text image editing, several tasks have been addressed like calligrams~\cite{xu2010structure,Maharik2011Micrography,Zou2016Legible} and handwriting generation~\cite{Haines2016My,Lian2016Automatic}. Lu~\textit{et~al}.~\cite{Lu2014DecoBrush} arranged and deformed pre-designed patterns along user-specified paths to synthesize decorative strokes. Handwriting style transfer~\cite{Lu2012HelpingHand} is accomplished using non-parametric samplings from a stroke library created by trained artists or parametric neural networks to learn stroke styles~\cite{Lian2016Automatic}.
However, most of these studies focus on text deformation. Much less has been done with respect to the fantastic text effects such as shadows, outlines, dancing flames (see Fig.~\ref{fig:supervsunsuper}), and soft clouds (see Fig.~\ref{fig:overview}).

To the best of our knowledge, the work of Yang~\textit{et al}.~\cite{Yang2017Awesome} is the only prior attempt at generating text effects. It solves the text stylization problem using a supervised texture transfer technique: a pair of registered raw text and its counterpart text effects are provided to calculate the distribution characteristics of the text effects, which guide the subsequent texture synthesis. In contrast, our framework automatically generates artistic typography, symbols and icons based on arbitrary source style images, without the input requirements as in~\cite{Yang2017Awesome}. Our method provides a more flexible and effective tool to create unique visual design artworks.

\footnotetext[1]{Unsplash (\url{https://unsplash.com/}) shares copyright-free photography from over 70,000 contributing photographers under the Unsplash license. We collect photos from Unsplash for use as style images and background images.}

\section{Problem Formulation and Framework}
\label{sec:overview}

We aim to automatically embed the target text-based binary shape in a background image with the style of a source reference image. To achieve this goal, we decompose the task into two subtasks: 1) Style transfer for migrating the style from source images to text-based binary shapes to design artistic shapes. 2) Layout design for seamlessly synthesizing artistic shapes in the background image to create visual design artwork such as posters and magazine covers.

Fig.~\ref{fig:overview} shows an overview of our algorithm. For the first subtask, we abstract a binary image from the source style image, adjust its contour and the outline of the target shape to narrow the structural difference between them. The adjusted results establish an effective mapping between the target binary image and the source style image. Then we are able to synthesize textures for the target shape. For the second subtask, we first seek the optimal layout of the target shape in the background image. Once the layout is determined, the shape is seamlessly synthesized into the background image under the constraint of the contextual information. The color statistics of the background image and the style image are optionally adjusted to ensure visual consistency.

\subsection{Style Transfer}
\label{sec:overview-task1}

The goal of text-based image style transfer is to stylize the target text-based binary image $T$ based on a style image $S'$. In previous text style transfer method~\cite{Yang2017Awesome}, distribution prior is a key factor to its success. However, this prior requires that $S'$ has highly structured textures with its non-stylized counterpart $S$ provided. By comparison, we solve a tougher unsupervised style transfer problem, where $S$ is not provided and $S'$ contains arbitrary textures. To meet these challenges, we propose to build a mapping between $T$ and $S'$ using a binary imagery $S$ of $S'$, and gradually narrow their visual discrepancy by structure and texture transfer. Moreover, a saliency cue is introduced for shape legibility.

In particular, instead of directly handling $S'$ and $T$, we first propose a two-stage abstraction method to abstract a binary imagery $S$ as a bridge based on the color features and contextual information of $S'$ (Section \ref{sec:foreground}). Since the textons in $S'$ and the glyphs in $T$ probably do not match, a legibility-preserving structure transfer algorithm is proposed to adjust the contours of $S$ and $T$ (Section \ref{sec:shape}). The resulting $\hat{S}$ and $\hat{T}$ share the same structural features and establish an effective mapping between $S'$ and $T$. Given $\hat{S}$ and $S'$, we are able to synthesize textures for $\hat{T}$ by objective function optimization (Section \ref{sec:texture}). In addition to the distribution term~\cite{Yang2017Awesome}, we further introduce a saliency term in our objective function, which guides our algorithm to stylize the interior of the target shape (white pixels in $\hat{T}$) to be of high saliency while the exterior (black pixels in $\hat{T}$) of low saliency. This principle enables the stylized shape to be highlighted from the background, thereby increasing its legibility.

\subsection{Layout Design}

The goal of context-aware layout design is to seamlessly synthesize $T$ into a background image $I$ with the style of $S'$. We formulate a optimization function to estimate the optimal embedding position $\hat{R}$ of $T$. And the proposed text-based image stylization is adjusted to incorporate contextual information of $I$ for seamless image synthesis.

In particular, the color statistics of $S'$ and $I$ are first adjusted to ensure color consistency (Section \ref{sec:color}). Then, we seek the optimal position of $T$ in the background image based on four cues of local variance, non-local saliency, coherence across images and visual aesthetics (Section \ref{sec:position}). Once the layout is determined, the background information around $T$ will be collected. Constrained by this contextual information, the target shape is seamlessly synthesized into the background image in an image inpainting manner (Section \ref{sec:embedding}).

\section{Text-Based Binary Image Style Transfer}
\label{sec:text_style_transfer}

\begin{figure}
  \centering
  \subfigure[$S'$]{
  \includegraphics[width=0.224\linewidth]{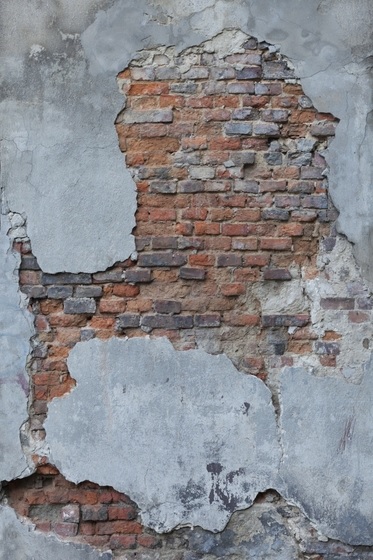}}
  \subfigure[$\bar{S}'$]{\label{fig:guidance_map1}
  \includegraphics[width=0.224\linewidth]{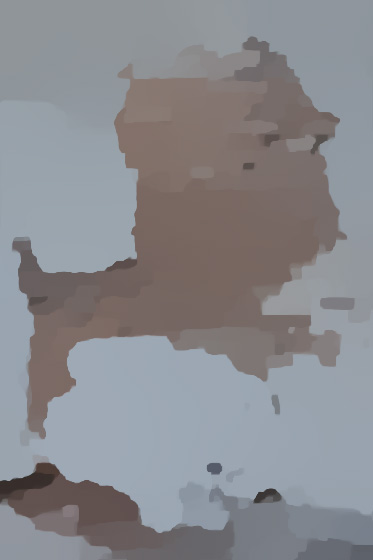}}
  \subfigure[\footnotesize{SP}$+\bar{S}'$]{
  \includegraphics[width=0.224\linewidth]{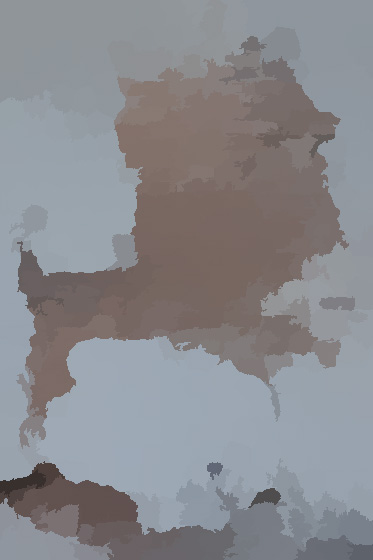}}
  \subfigure[$S$]{
  \includegraphics[width=0.224\linewidth]{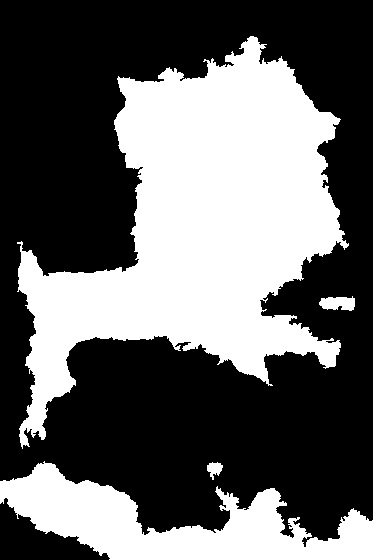}}\vspace{-1mm}
  \subfigure[\footnotesize{raw cluster}]{\label{fig:guidance_map2}
  \includegraphics[width=0.224\linewidth]{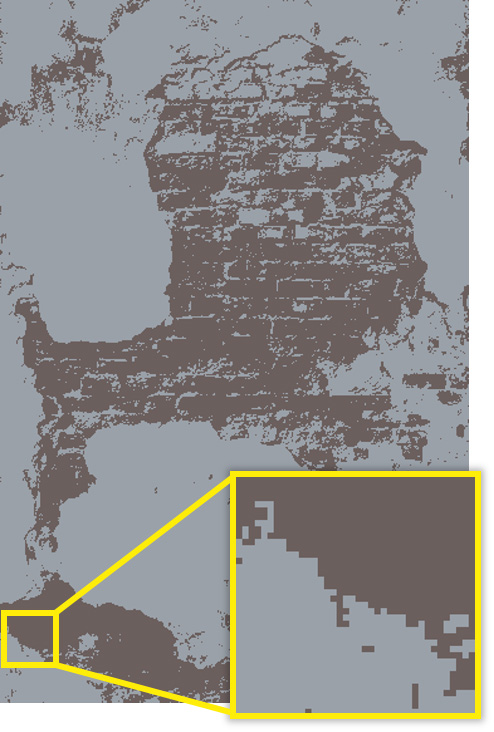}}
  \subfigure[\footnotesize{one-stage}]{
  \includegraphics[width=0.224\linewidth]{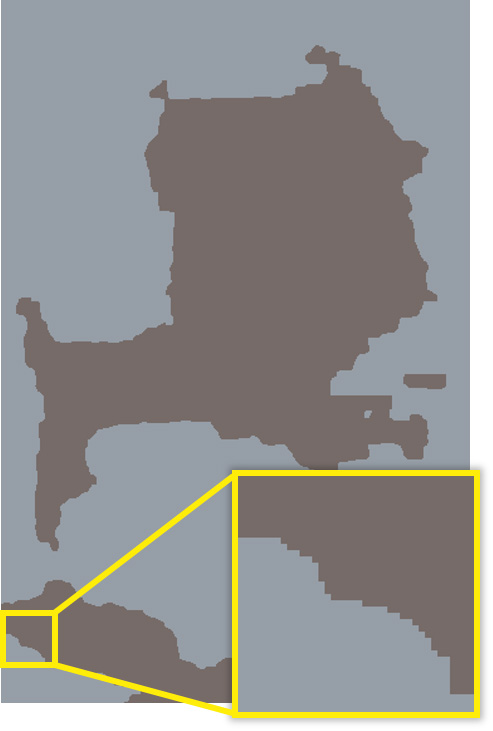}}
  \subfigure[\footnotesize{two-stage}]{
  \includegraphics[width=0.224\linewidth]{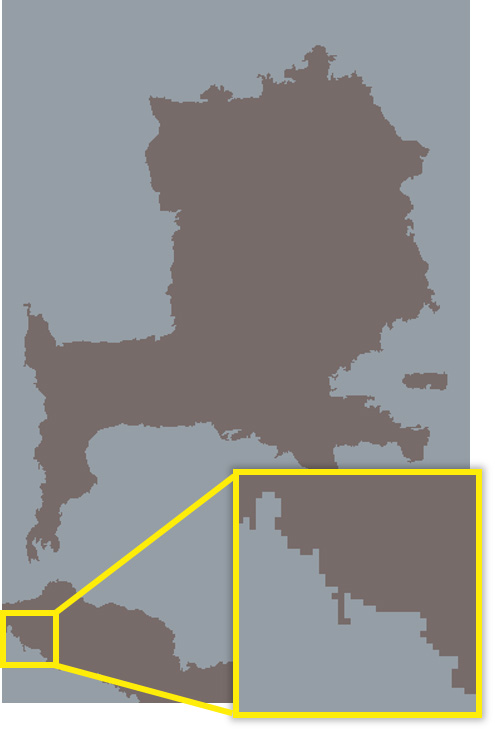}}
  \subfigure[\footnotesize{result by~\cite{lockerman2016multi}}]{\label{fig:guidance_map3}
  \includegraphics[width=0.224\linewidth]{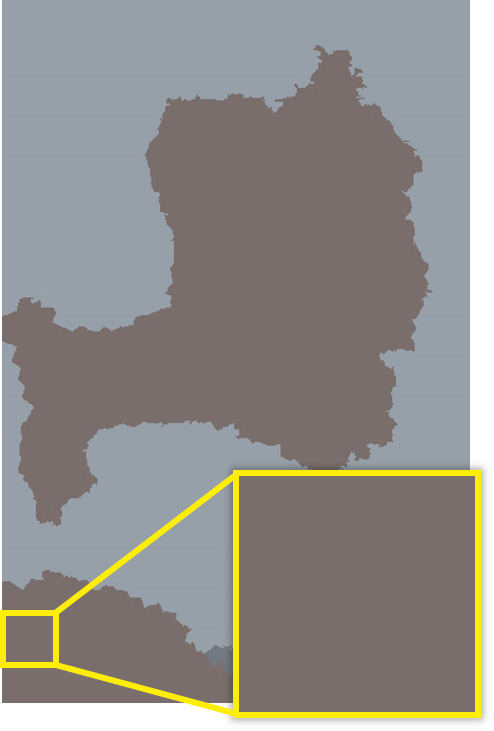}}
  \caption{Guidance map extraction. Our two-stage abstraction method generates guidance maps that well match the texture contours in $S'$. (a) Input $S'$. (b) Rough structure abstraction $\bar{S}'$. (c) Super pixels colored by their mean pixel values in (b). (d) Extracted guidance map $S$. (e)-(g) K-means clustering results of (a)-(c), respectively. (h) Result by multi-scale label-map extraction~\cite{lockerman2016multi}. Cropped regions are zoomed for better comparison.}\label{fig:guidance_map}
  \vspace{-2mm}
\end{figure}

\subsection{Guidance Map Extraction}
\label{sec:foreground}

The perception of texture is a process of acquiring abstract imagery, which enables us to see concrete images from the disordered (such as clouds). This inspires us to follow human's abstraction of the texture information to extract the binary imagery $S$ from the source image $S'$. $S$ serves as a guidance map, where white pixels indicate the reference region for the shape interior (foreground) and black pixels for the shape exterior (background). The boundary of foreground and background depicts the morphological characteristics of the textures in $S'$. We propose a simple yet effective two-stage method to abstract the texture into the foreground and the background with the help of texture removal technologies.

In particular, we use the Relative Total Variation~(RTV)~\cite{xu2012structure} to remove the color variance inside the texture, and obtain a rough structure abstraction $\bar{S}'$. However, texture contours are also smoothed in $\bar{S}'$ (see Fig.~\ref{fig:guidance_map}(b)(f)). Hence, we put forward a two-stage abstraction method. In the first stage, pixels in $S'$ are abstracted as fine-grained super pixels~\cite{achanta2012slic} to precisely match the texture contour. Each super pixel uses its mean pixel values in $\bar{S}'$ as its feature vector to avoid the texture variance. In the second stage, the super pixels are further abstracted as the coarse-grained foreground and background via $K$-means clustering ($K=2$).  Fig.~\ref{fig:guidance_map} shows an example where our two-stage method generates accurate abstract imagery of the plaster wall. In this example, our result has more details at the boundary than the one-stage method, and fewer errors than the state-of-the-art label-map extraction method~\cite{lockerman2016multi}~(see the zoomed region in Fig.~\ref{fig:guidance_map}(h)).

Finally, we use the saliency as a criterion to determine the foreground and background of the image. Pixel saliency in $S'$ is detected~\cite{zhang2013saliency} and the cluster with higher mean pixel saliency is set as the foreground. Compared with the commonly used brightness criterion in artistic thresholding methods~\cite{Mould2008Stylized,Xu2008Artistic} to retrieve artistic binary images, our criterion helps the foreground text find salient textures.

\begin{figure}
  \centering
  \subfigure{\label{fig:shape_transfer1}
  \includegraphics[width=0.25\linewidth]{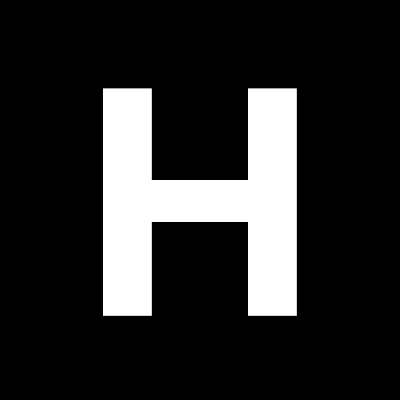}}
  \subfigure{
  \includegraphics[width=0.336\linewidth]{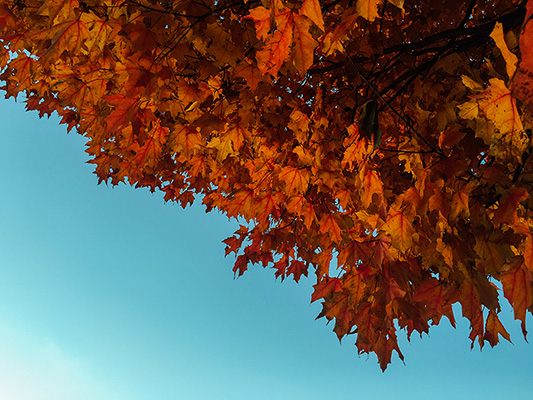}}
  \subfigure{
  \includegraphics[width=0.336\linewidth]{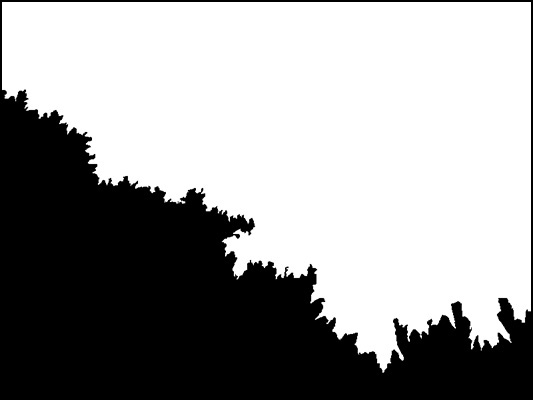}}
  \addtocounter{subfigure}{-3}
  \subfigure[]{\label{fig:shape_transfer2}
  \includegraphics[width=0.224\linewidth]{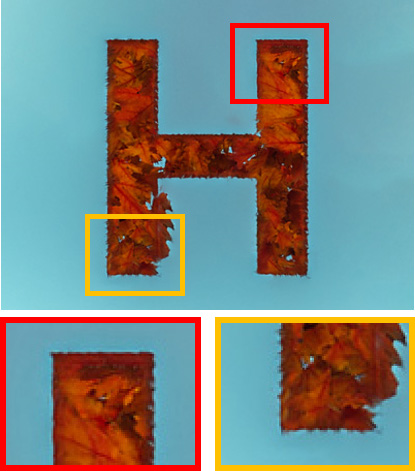}}
  \subfigure[]{
  \includegraphics[width=0.224\linewidth]{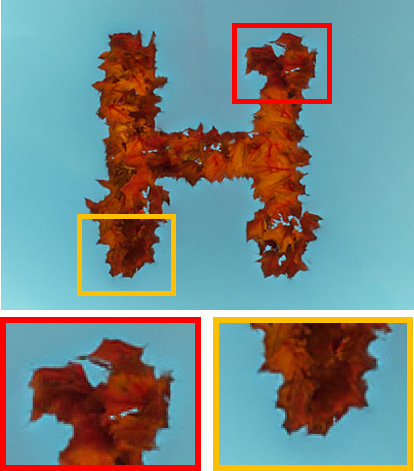}}
  \subfigure[]{
  \includegraphics[width=0.224\linewidth]{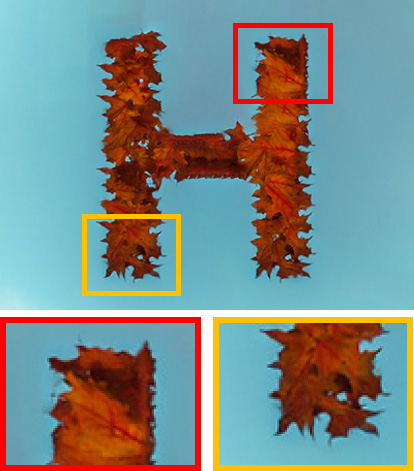}}
  \subfigure[]{\label{fig:shape_transfer3}
  \includegraphics[width=0.224\linewidth]{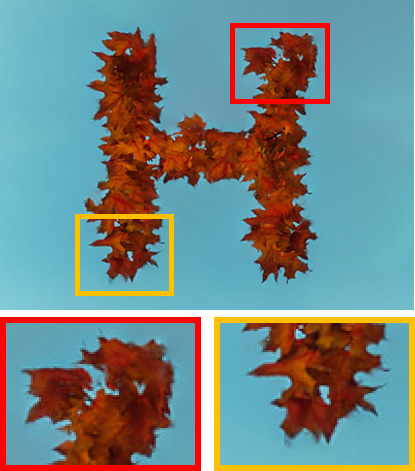}}
  \caption{Benefits of bidirectional structure transfer. The forward transfer simulates the distribution of leaves along the shape boundary, while the backward transfer generates the fine details of each leaf shape. Their combination creates vivid leaf-like typography.
  The first row: from left to right, $T$, $S'$ and $S$. The second row: the text stylization results using (a) original $T+S$, (b) forward transfer $\hat{T}+S$,  (c) backward transfer $T+\hat{S}$, and (d) bidirectional transfer $\hat{T}+\hat{S}$. \textit{Image credits: Unsplash users Aaron Burden.}
  }\label{fig:shape_transfer}
\end{figure}

\subsection{Structure Transfer}
\label{sec:shape}

\begin{figure}
  \centering
  \subfigure{
  \includegraphics[width=0.174\linewidth]{figures/shap-lv-0.jpg}}
  \subfigure{
  \includegraphics[width=0.174\linewidth]{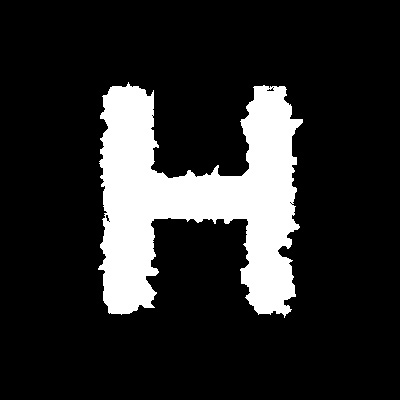}}
  \subfigure{
  \includegraphics[width=0.174\linewidth]{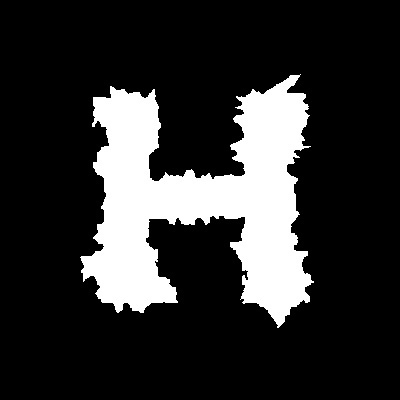}}
  \subfigure{
  \includegraphics[width=0.174\linewidth]{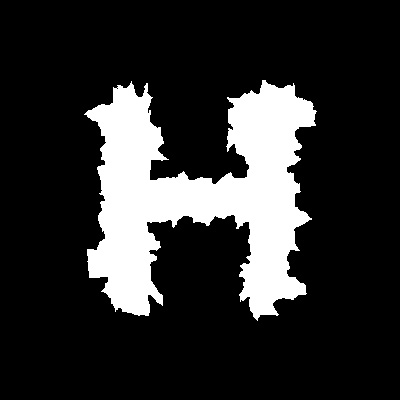}}
  \subfigure{
  \includegraphics[width=0.174\linewidth]{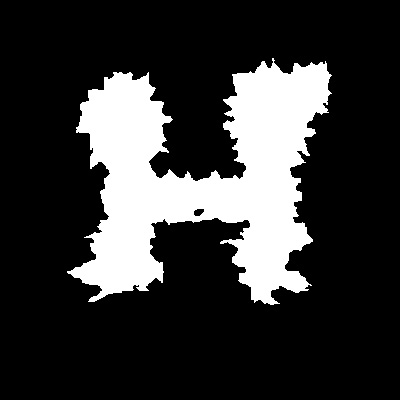}}\hfill
  \addtocounter{subfigure}{-5}
  \subfigure[$L=0$]{
  \includegraphics[width=0.174\linewidth]{figures/shap-lv-0.jpg}}
  \subfigure[$L=3$]{
  \includegraphics[width=0.174\linewidth]{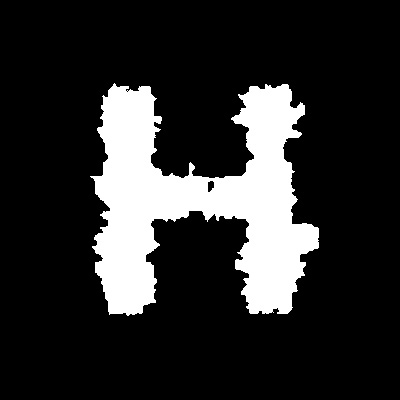}}
  \subfigure[$L=7$]{
  \includegraphics[width=0.174\linewidth]{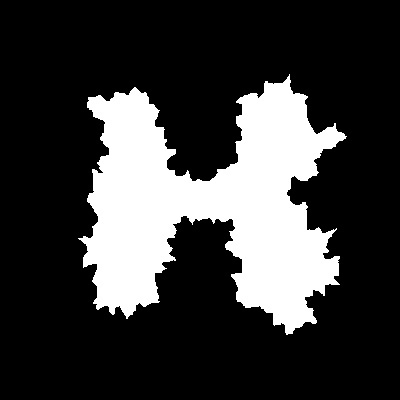}}
  \subfigure[$L=11$]{
  \includegraphics[width=0.174\linewidth]{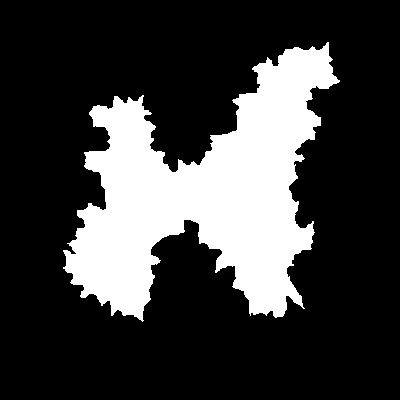}}
  \subfigure[$L=15$]{
  \includegraphics[width=0.174\linewidth]{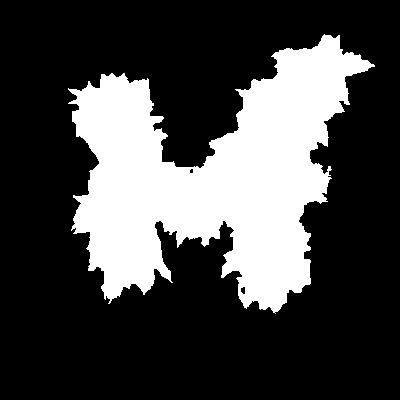}}
  \caption{Effects of our stroke trunk protection mechanism and the number $L$ of image pyramid layers. Top row: our legibility-preserving structure transfer result $\hat{T}^{(0)}$. Bottom row: structure transfer result $\hat{T}^{(0)}$ without our stroke trunk protection mechanism. As $L$ increases, the deformation degree also increases. Without the stroke trunk protection mechanism, the shape legibility is severely reduced. `H' is deformed like `X' in (d) and `M' in (e) in the bottom row. In this example, we use $S$ in Fig.~\ref{fig:shape_transfer} as the reference.}\label{fig:shape_level}
\end{figure}

Directly using $S$ extracted in Section \ref{sec:foreground} and the input $T$ for style transfer results in unnatural texture boundaries as shown in Fig.~\ref{fig:shape_transfer}(a). A potential solution could be employing the shape synthesis technique~\cite{rosenberger2009layered} to minimize structural inconsistencies between $S$ and $T$. In Layered Shape Synthesis (LSS)~\cite{rosenberger2009layered}, shapes are represented as a collection of boundary patches at multiple resolution, and the style of a shape is transferred onto another by optimizing a bidirectional similarity function. However, in our application such an approach does not consider the legibility, and the shape will become illegible after adjustment as shown in the second row of Fig.~\ref{fig:shape_level}. Hence we incorporate stroke trunk protection mechanism into LSS and propose a legibility-preserving structure transfer method.

\begin{figure}
  \centering
  \subfigure{
  \includegraphics[width=0.98\linewidth]{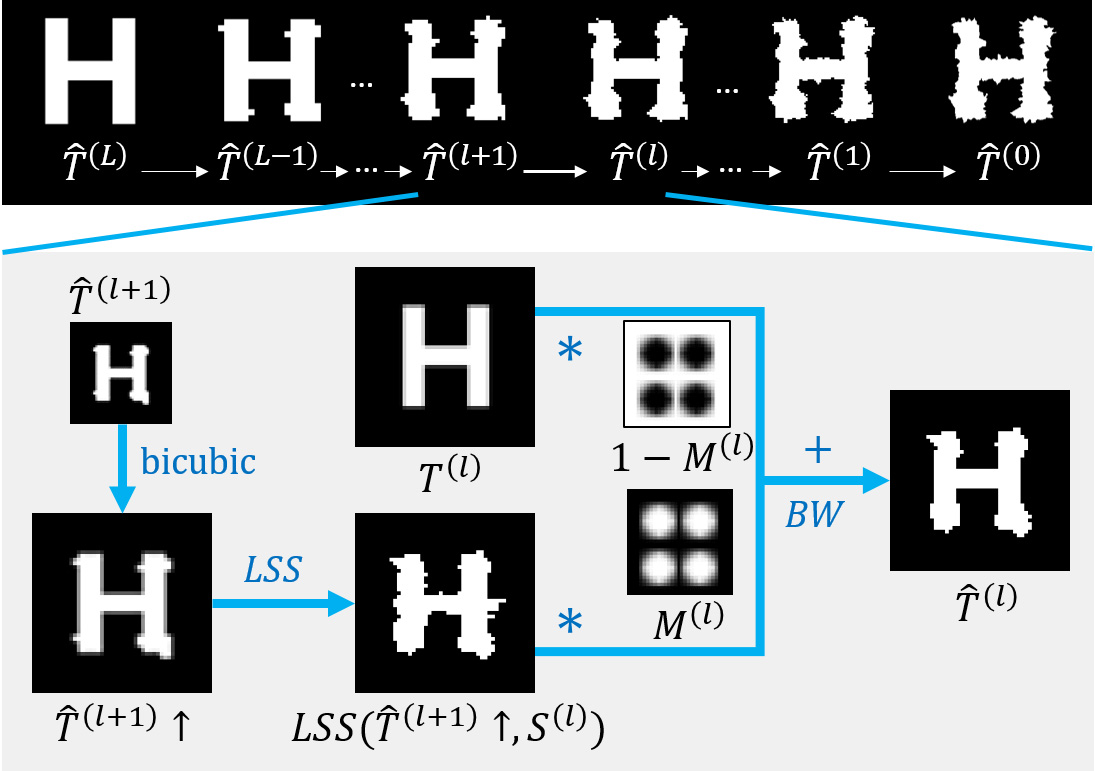}}
  \caption{Preserving the stroke trunks by weighted combination of the trunk region from $T^{(l)}$ and the stroke end region from the shape synthesis result.}\label{fig:shape_pipeline}
\end{figure}

The main idea is to adjust the shape of the stroke ends while preserving the shape of the stroke trunk, because the legibility of a glyph is mostly determined by the shape of its trunk. Toward this, we extract the skeleton from $T$ and detect the stroke end as a circular region centered at the endpoint of the skeleton. For each resolution level $l$, we generate a mask $M^{(l)}$ indicating the stroke end regions as shown in Fig.~\ref{fig:shape_pipeline}. At the top level $l=L$, the radius of the circular region is set to the average radius of the stroke calculated by the method of \cite{Yang2017Awesome}. The radius increases linearly as the resolution increases. And at the bottom level $l=0$ (original resolution), it is set to just cover the entire shape. Let $T^{(l)}$, $S^{(l)}$ and $\hat{T}^{(l)}$ denote the downsampled $T$, downsampled $S$ and the legibility-preserving structure transfer result at level $l$ respectively. Given $M^{(l)}$, $T^{(l)}$, $S^{(l)}$ and $\hat{T}^{(l+1)}$, we calculate $\hat{T}^{(l)}$ by
\begin{equation}
\label{eq:shape_transfer}
\hat{T}^{(l)}=BW\Big(M^{(l)}\ast LSS\big(\hat{T}^{(l+1)}\uparrow,S^{(l)}\big)+(1-M^{(l)})\ast T^{(l)}\Big),
\end{equation}
where $\ast$ is the element-wise multiplication operator and $\uparrow$ is the bicubic upsampling operator. $LSS(T,S)$ is the shape synthesis result of $T$ given $S$ as the shape reference by LSS, and $BW(\cdot)$ is the binarization operation with threshold $0.5$. The pipeline of the proposed structure transfer is visualized in Fig.~\ref{fig:shape_pipeline}. In our implementation, the image resolution at the top level $L$ is fixed. Therefore the deformation degree is solely controlled by $L$. We show in Fig.~\ref{fig:shape_level} that our stroke trunk protection mechanism effectively balances structural consistency with shape legibility even under very large $L$.

\begin{figure}[t]
\vspace{-3mm}
  \centering
  \subfigure[$S'$]{
  \includegraphics[width=0.36\linewidth]{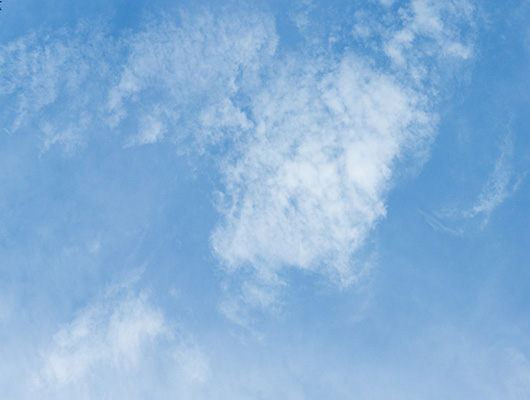}}
  \subfigure[Saliency map of $S'$]{
  \includegraphics[width=0.44\linewidth]{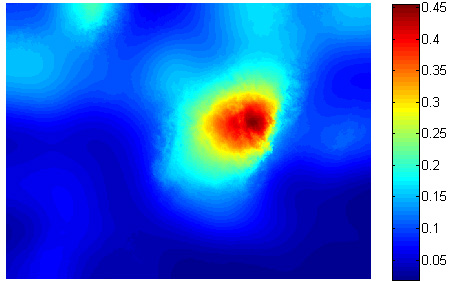}}
  \subfigure[$\lambda_3=0.00$]{
  \includegraphics[width=0.304\linewidth]{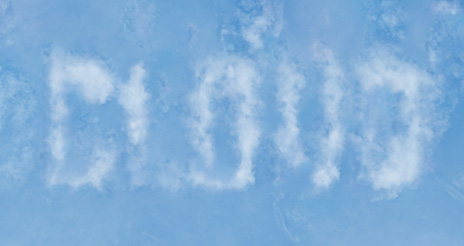}}
  \subfigure[$\lambda_3=0.01$]{
  \includegraphics[width=0.304\linewidth]{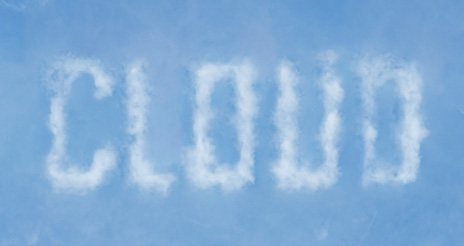}}
  \subfigure[$\lambda_3=0.05$]{
  \includegraphics[width=0.304\linewidth]{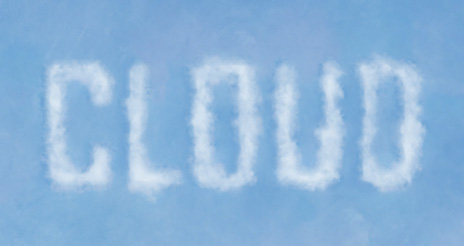}}
  \caption{Effects of the saliency term in texture transfer. The saliency term makes the foreground character more prominent and the background cleaner, thereby enhancing the legibility.
  \textit{Image credits: Unsplash users Ashim D'Silva.}}\label{fig:saliency_transfer}
\end{figure}

For characters without stroke ends (\textit{e.g.} ``O''), our method automatically reverts to the baseline LSS algorithm. We use an eclectic strategy to keep these characters consistent with the average degree of adjustment of characters with stroke ends. Specifically, during our $L$-level hierarchical shape transfer, these characters are masked out during the first $L/2$ levels, and then are adjusted using LSS in the following $L/2$ levels.

In addition, we propose a bidirectional structure transfer (Fig.~\ref{fig:shape_transfer}(a)-(d)) to further enhance the shape consistency, where a backward transfer is added after the aforementioned forward transfer. The backward transfer migrates the structural style of the forward transfer result $\hat{T}^{0}$ back to $S$ to obtain $\hat{S}^{0}$ using the original LSS algorithm. The results $\hat{T}^{(0)}$ and $\hat{S}^{(0)}$ will be used as guidance for texture transfer. For simplicity, we will omit the superscripts in the following.

\subsection{Texture Transfer}
\label{sec:texture}

In our scenarios, $S'$ is not well-structured text effects, and thus the distribution prior used in \cite{Yang2017Awesome} to ensure shape legibility takes limited effect. We introduce a saliency cue for compensation. We augment the texture synthesis objective function in \cite{Yang2017Awesome} with the proposed saliency term as follows,
\begin{equation}
\label{eq:objective_function}
\min_{q}\sum_{p}E_a(p,q)+\lambda_1E_d(p,q)+\lambda_2E_p(p,q)+\lambda_3E_s(p,q),
\end{equation}
where $p$ is the center position of a target patch in $\hat{T}$ and $T'$, $q$ is the center position of the corresponding source patch in $\hat{S}$ and $S'$. The four terms $E_a$, $E_d$, $E_p$ and $E_s$ are the appearance, distribution, psycho-visual and saliency terms, respectively, weighted by $\lambda$s. $E_a$ and $E_d$ constrain the similarity of local texture pattern and global texture distribution, respectively. $E_p$ penalizes texture over-repetitiveness for naturalness. We refer to~\cite{Yang2017Awesome} for details of the first three terms. For the distribution term $E_d$, we truncate the distance maps of $\hat{S}$ and $\hat{T}$ to a range of $[0.5, 2]$ where distance $1$ corresponds to the shape boundaries. By doing so, we relieve the distribution constraint for pixels far away from the shape boundary. And these pixels are mainly controlled by our saliency cue,
\begin{equation}\label{eq:saliency}
E_s(p,q)=\left\{
\begin{aligned}
& W(p)\cdot Sal(q),\text{ if } T(p)=0\\
& W(p)\cdot(1-Sal(q)),\text{ if } T(p)=1\\
\end{aligned}
\right.
\end{equation}
where $Sal(q)$ is the saliency at pixel $q$ in $S'$. $W(p)=1-\exp(-dist(p)^2/2\sigma_1^2)/2\pi\sigma^2_1$ is the gaussian weight with $dist(p)$, the distance of $p$ to the shape boundary. The saliency term encourages pixels inside the shape to find salient textures for synthesis and keeps the background less salient. We show in Fig.~\ref{fig:saliency_transfer} that a higher weight of our saliency term makes the stylized shape more prominent.

\begin{figure}
  \centering
  \subfigure[Input $T$, $S'$ and $I$]{
  \includegraphics[width=0.58\linewidth]{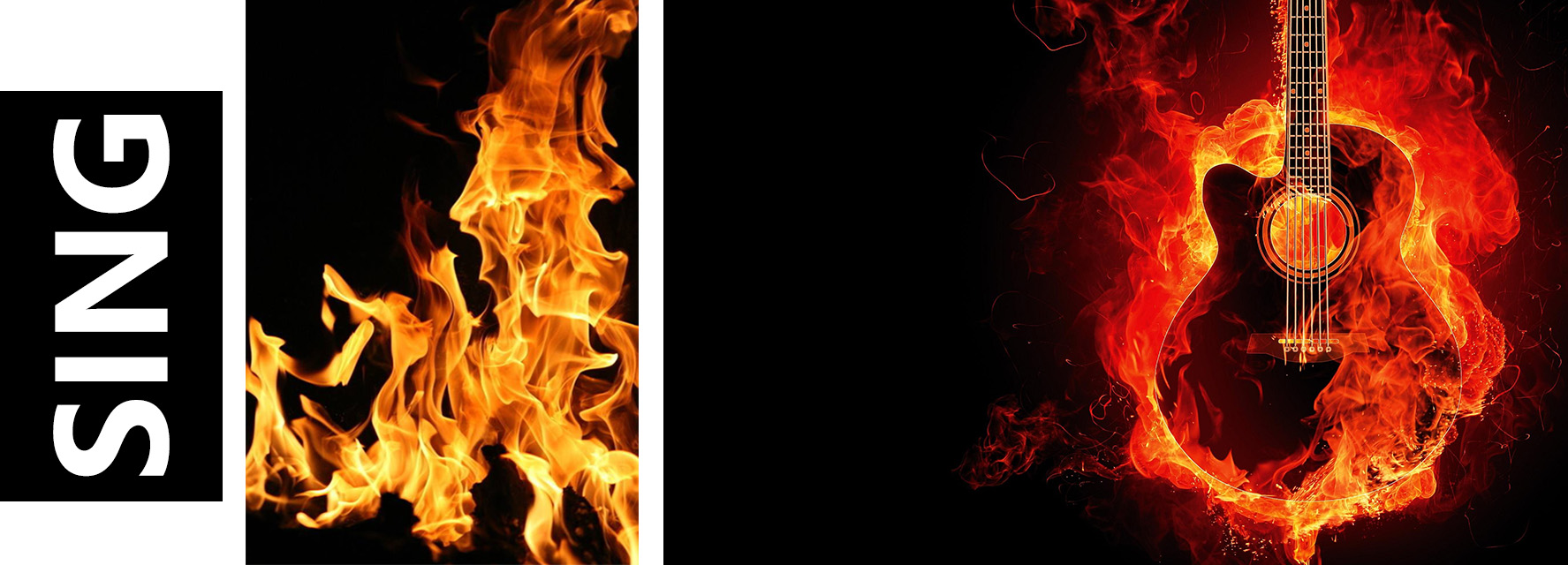}}
  \subfigure[Total placement cost]{
  \includegraphics[width=0.366\linewidth]{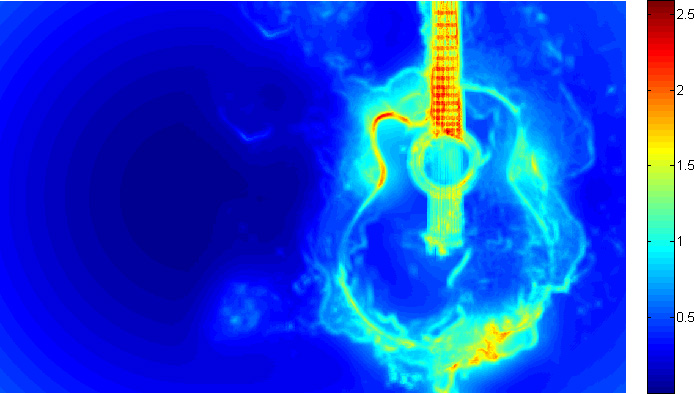}}
  \subfigure[Variance]{
  \includegraphics[width=0.224\linewidth]{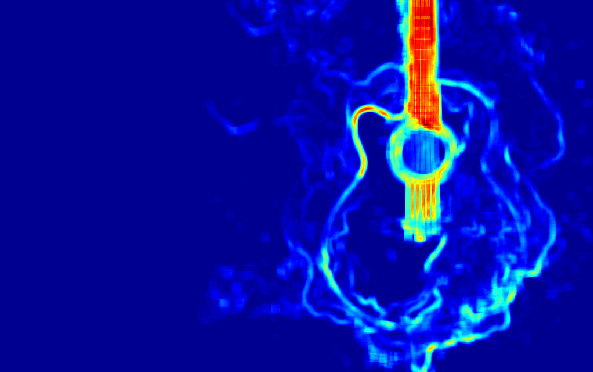}}
  \subfigure[Saliency]{
  \includegraphics[width=0.224\linewidth]{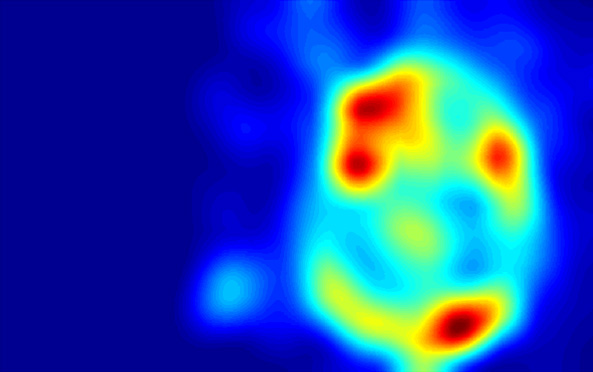}}
  \subfigure[Coherence]{
  \includegraphics[width=0.224\linewidth]{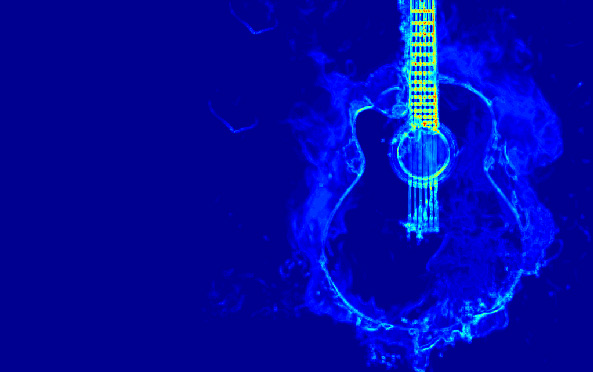}}
  \subfigure[Aesthetics]{
  \includegraphics[width=0.224\linewidth]{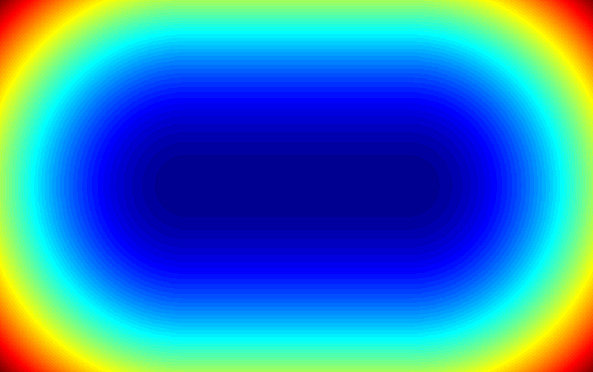}}
  \subfigure[Result without the aesthetics cost]{\label{fig:position_estimation1}
  \includegraphics[width=0.47\linewidth]{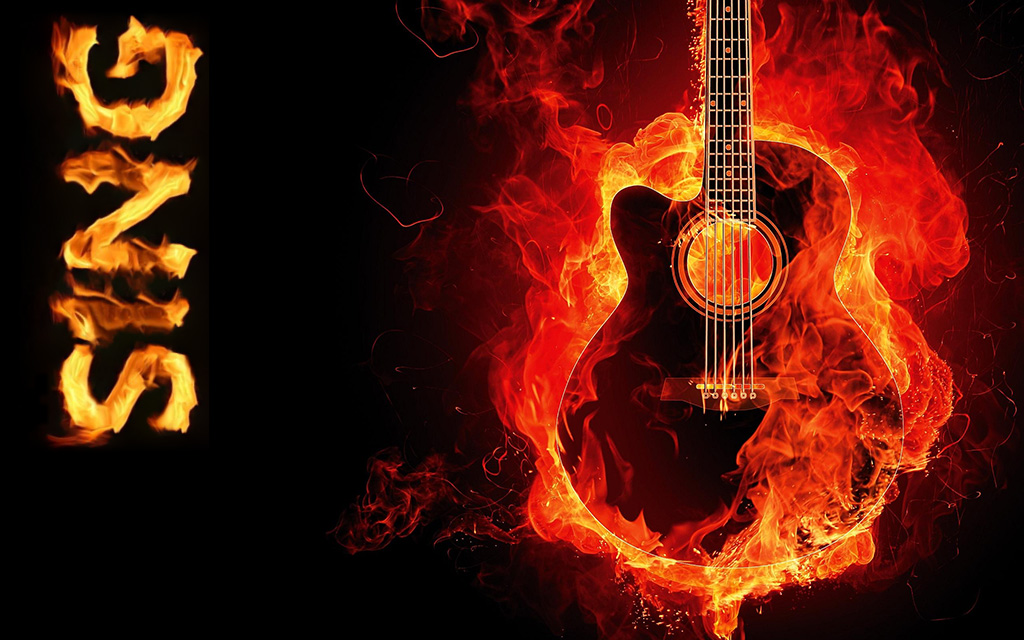}}
  \subfigure[Result with the aesthetics cost]{
  \includegraphics[width=0.47\linewidth]{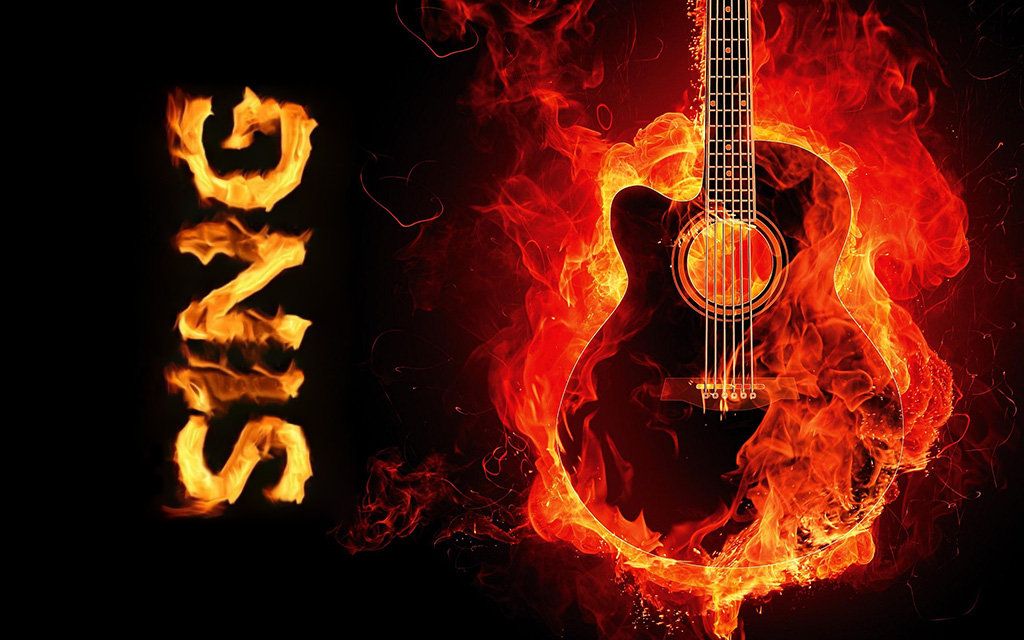}}
  \caption{Image layout is determined by jointly considering a local variance cost,  a non-local saliency cost, a coherence cost and an aesthetics cost.
  \textit{Image credits: Unsplash users Dark Rider.}}\label{fig:position_estimation}
\end{figure}

Similar to \cite{Yang2017Awesome}, we take the iterative coarse-to-fine matching and voting steps as in \cite{Space-time}. In the matching step, PatchMatch algorithm \cite{PatchMatch} is adopted to solve (\ref{eq:objective_function}).

\section{Context-Aware Layout Design}
\label{sec:text_embedding}

\subsection{Color Transfer}
\label{sec:color}

Obvious color discontinuities may appear for style images $S'$ that have a different color from the background $I$. Therefore we employ color transfer technology. Here we use a linear method introduced by Image Analogies color transfer~\cite{hertzmann2001algorithms}. This technique
estimates a color affine transformation matrix and a bias vector which match the target mean and standard deviation of the color feature with the source ones. In general, color transfer in a local manner is more robust than the global method. Hence, we employ the perception-based color clustering technique~\cite{chang2006example} to divide pixels into eleven color categories. The linear color transfer is performed within corresponding categories between $S'$ and $I$. More sophisticated methods or user interactions could be optionally employed to further improve the color transfer result.

\subsection{Position Estimation}
\label{sec:position}

In order to synthesize the target shape seamlessly into the background image, the image layout should be properly determined. In the literature, similar problems in cartography are studied to place text labels on maps~\cite{Christensen1995An}. Viewed as an optimization problem, they only consider the overlap between labels. In this paper, both the seamlessness and aesthetics of text placement are taken into account.
Specifically, we formulate a cost minimization problem for context-aware position estimation by considering the cost of each pixel $x$ of $I$ in four aspects,
\begin{equation}
\label{eq:position_function}
\hat{R}=\arg\min_{R}\sum_{x\in R}U_v(x)+U_s(x)+U_c(x)+\lambda_4U_a(x),
\end{equation}
where $R$ is a rectangular area of the same size as $T$, indicating the embedding position. The position is estimated by searching an $\hat{R}$ where pixels have the minimum total costs. $U_v$ and $U_s$ are local variance and non-local saliency costs, concerning the background image $I$ itself, and $U_c$ is a coherence cost measuring the coherence between $I$ and $S'$. In addition, $U_a$ is the aesthetics cost for subjective evaluation, weighted by $\lambda_4$. Here all terms are normalized independently. We use equal weights for the first three terms, and a lower weight $\lambda_4=0.5$ for the aesthetics term.

We consider the local and non-local cues of $I$. First, we seek flat regions for seamless embedding by using $U_v(x)$ as the intensity variance within a local patch centered at $x$. Then, a saliency cost $U_s(x)=Sal(x)$ which prefers non-salient regions is introduced. These two internal terms preclude our method from overlaying important objects in the background image with the target shape.

In addition, we use the mutual cost $U_c$ to measure the texture consistency between $I$ and $S'$. More specifically, $U_c(x)$ is obtained by calculating the $L2$ distance between the patch centered at $x$ in $I$ and its best matched patch in $S'$.

\begin{figure}[t]
  \centering
  \subfigure{
  \includegraphics[width=0.98\linewidth]{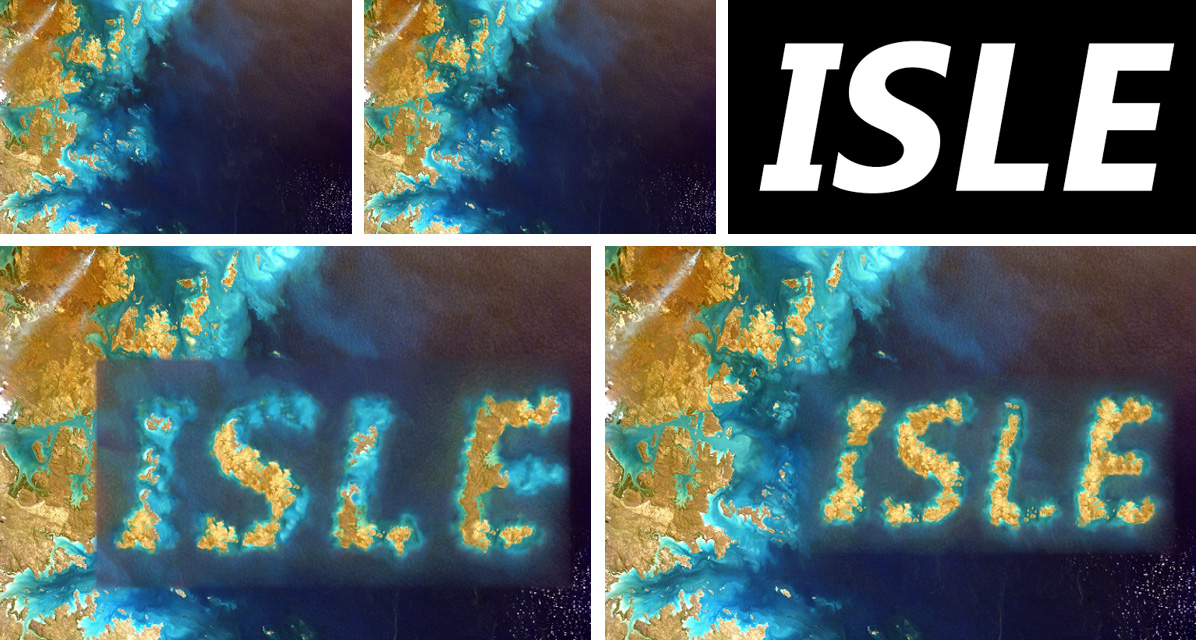}}
  \caption{Text scale in position estimation. Top row: Three images from left to right are $I$, $S'$ and $T$, respectively. Bottom row: the result image layout using $T$ of the original size (left) and the result image layout using $T$ of the optimal size (right). \textit{Image credits: Unsplash user NASA.}}\label{fig:pos-scale}\vspace{-2mm}
\end{figure}

So far, both the internal and mutual cues are modeled. However, a model that considers only the seamlessness may find unimportant image corners for the target shape, which is not ideal for aesthetics as shown in Fig.~\ref{fig:position_estimation}(g). Hence, we also model the centrality of the shape by $U_a$
\begin{equation}
U_a(x)=1-\exp\big(-dist(x)^2/2\sigma_2^2\big),
\end{equation}
where $dist(x)$ is the offset of $x$ to the image center, and $\sigma_2$ is set to the length of the short side of $I$. Fig.~\ref{fig:position_estimation} visualizes these four costs, which jointly determine the ideal image layout.

As for the minimization of (\ref{eq:position_function}), we use the box filter to effectively solve the total costs for every valid $R$ throughout $I$, and choose the minimum one.

We further consider the scales and rotations of text. In addition, when the target image contains multiple shapes (which is quite common for text), we investigate the placement of each individual shape rather than considering them as a whole, which greatly enhances the design flexibility.

\begin{figure}[t]
  \centering
  \subfigure{
  \includegraphics[width=0.98\linewidth]{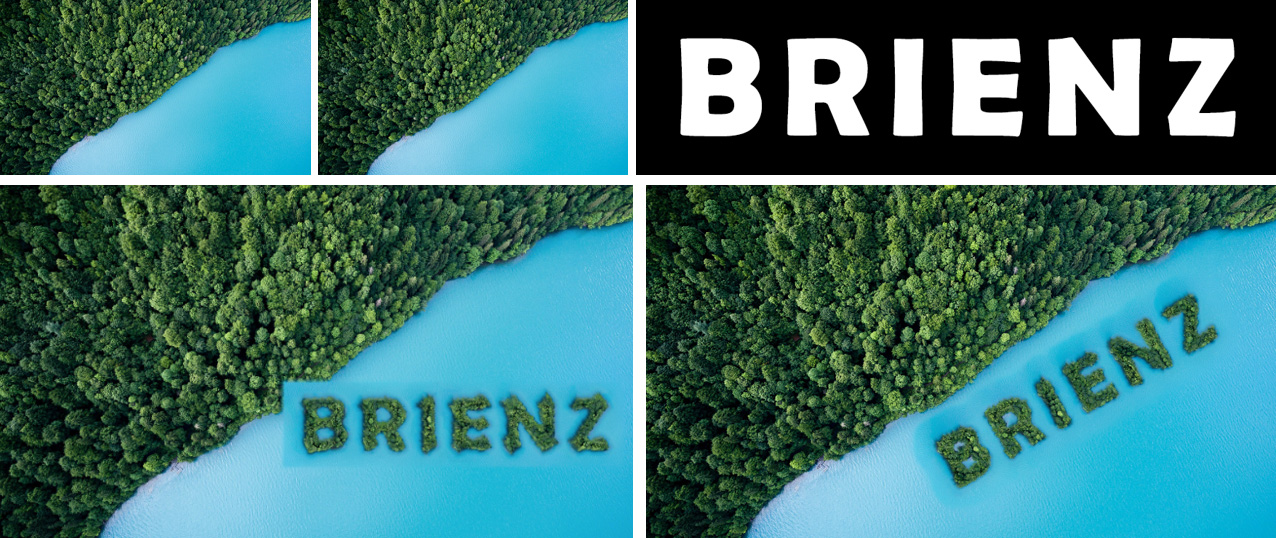}}
  \caption{Text rotation in position estimation. Top row: Three images from left to right are $I$, $S'$ and $T$, respectively. Bottom row: the result image layouts without text rotation (left) and with text rotation (right). \textit{Image credits: Unsplash user Andreas Gucklhor.}}\label{fig:pos-rotation}
\end{figure}

\begin{figure}[t]
  \centering
  \subfigure{
  \includegraphics[width=0.98\linewidth]{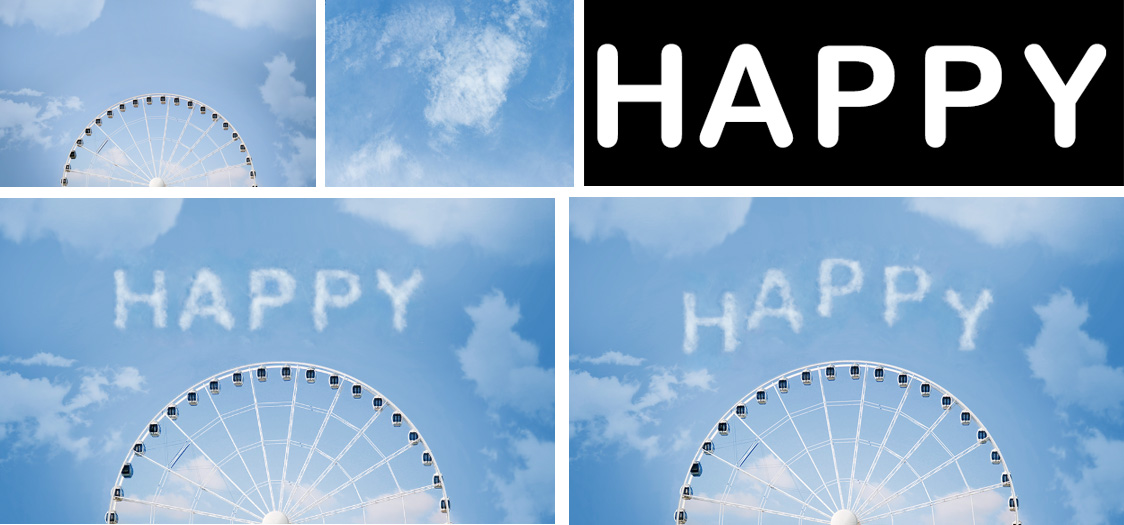}}
  \caption{Multi-shape placement in position estimation. Top row: Three images from left to right are $I$, $S'$ and $T$, respectively. Bottom row: the result image layouts without layout refinement (left) and with layout refinement (right).}\label{fig:pos-multi}
\end{figure}

\textbf{Text scale}: Our method can be easily extended to handle different scales. During the box filtering, we enumerate the size of the box and then find the global minimum penalty point throughout the space and scale. Specifically, we enumerate a scaling factor $\psi_s$ over a range of $[0.8, 1.2]$ in steps of $0.1$. Then the text box $R$ is zoomed in or out based on $\psi_s$ to obtain $\psi_s(R)$. Finally, the optimal embedding position and scale can be detected by
\begin{equation}
\label{eq:position_function2}
\hat{R},\hat{\psi_s}=\arg\min_{R,\psi_s}\sum_{x\in \psi_s(R)}\frac{U_v(x)+U_s(x)+U_c(x)+\lambda_4U_a(x)}{|\psi_s(R)|}.
\end{equation}
Fig.~\ref{fig:pos-scale} shows an example where the target image $T$ is originally too large and is automatically adjusted by the proposed method so that it could be seamlessly embedded into the background.

\textbf{Text rotation}: Similar to the text scale, we enumerate the rotation angle $\psi_r$ over a range of $[-\pi/6, \pi/6]$ in steps of $\pi/60$, and find the global minimum penalty point in the entire space and angle. To use the box filter for fast solution, instead of rotating $R$, we choose to rotate the cost map $\mathbf{U}=U_v(x)+U_s(x)+U_c(x)+\lambda_4U_a(x)$ by $-\psi_r$, and perform box filter on the rotated $\mathbf{U}$, followed by minimum point detection. Fig.~\ref{fig:pos-rotation} shows an example where the target image $T$ is automatically rotated to match the direction of the coastline.

\begin{figure*}[t]
  \centering
  \includegraphics[width=0.97\linewidth]{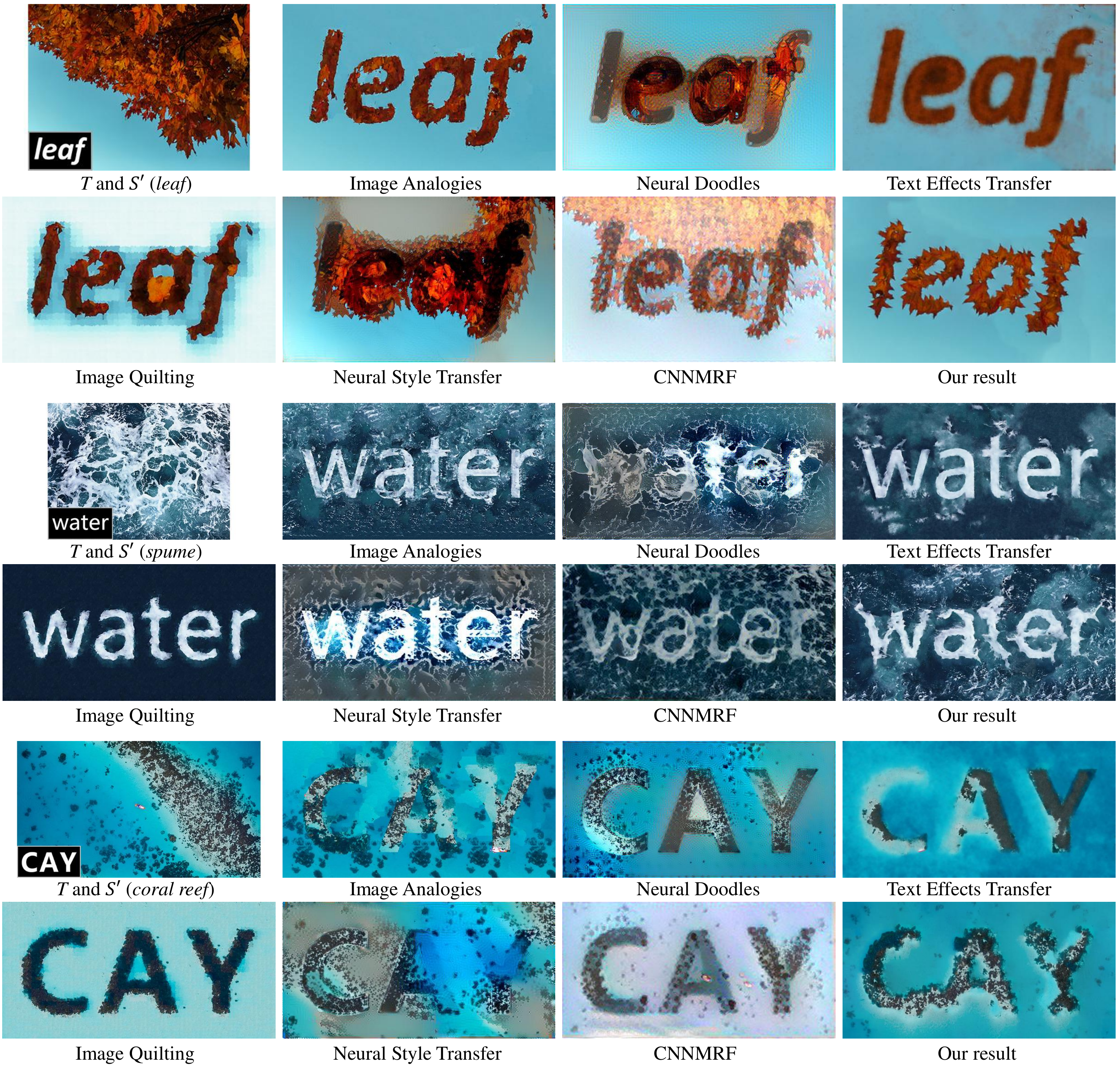}
  \caption{Visual comparison of text stylization. For each result group, the first one is the input source style and target text. Other images are results by supervised Image Analogies~\cite{Hertzmann2001Image}, Neural Doodles~\cite{Champandard2016Semantic}, Text Effects Transfer~\cite{Yang2017Awesome} (the upper row) and unsupervised Image Quilting~\cite{Efros2001Image}, Neural Style~\cite{gatys2016image}, CNNMRF~\cite{Li2016Combining}, our methods (the lower row). For supervised methods, the structure guidance map extracted by our method is directly given as input. More examples can be found in the supplementary material.
  \textit{Image credits: Unsplash users  Aaron Burden, Cassie Matias, Ishan@seefromthesky.}}\label{fig:experiment1}
\end{figure*}

\textbf{Multiple shapes}: To deal with multiple shapes, we first view them as a whole and optimize (\ref{eq:position_function}) to search an initial position and then refine their layouts separately. In each refinement step, every shape searches for the location with the lowest cost within its small spatial neighborhood to update its original position. After several steps, all the shapes converge to their respective optimal positions. In order to prevent the shapes from overlapping, the search space is limited to ensure the distance between adjacent shapes is not less than their initial distance. Fig.~\ref{fig:pos-multi} shows that after layout refinement, the characters on the left and right sides are adjusted to a more central position in the vertical direction, making the overall text layout better match the shape of the Ferris wheel.

It is worth noting that the above three extensions can be combined with each other to provide users with more flexible layout options.

\subsection{Shape Embedding}
\label{sec:embedding}

Once the layout is determined, we synthesize the target shape into the background image in an image inpainting manner. Image inpainting technologies~\cite{bertalmio2000image,Criminisi04region,Meur13Hierarchical} have long been investigated in image processing literature to fill the unknown parts of an image. Similarly our problem sets $\hat{R}$ as the unknown region of $I$, and we aim to fill it with the textures of $S'$ under the structure guidance of $\hat{T}$ and $\hat{S}$. We first enlarge $\hat{R}$ by expanding its boundary by $32$ pixels. Let the augmented frame-like region be denoted as $\hat{R}^{+}$, and the pixel values of $I$ in $\hat{R}^{+}$ provide contextual information for the texture transfer. Throughout the coarse-to-fine texture transfer process described in Section \ref{sec:texture}, each voting step is followed by replacing the pixel values of $T'$ in $\hat{R}^{+}$ with the contextual information $I(\hat{R}^{+})$. This manner will enforce a strong boundary constraint to ensure a seamless transition at the boundary.

\section{Experimental Results and Analysis}
\label{sec:experiment}

\begin{figure}
  \centering
  \subfigure[\footnotesize{Target text $T$}]{
  \includegraphics[width=0.48\linewidth]{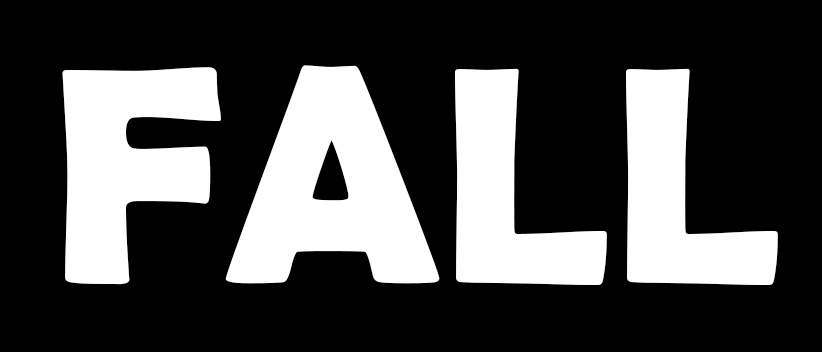}}
  \subfigure[\footnotesize{Adjusted text $\hat{T}$}]{
  \includegraphics[width=0.48\linewidth]{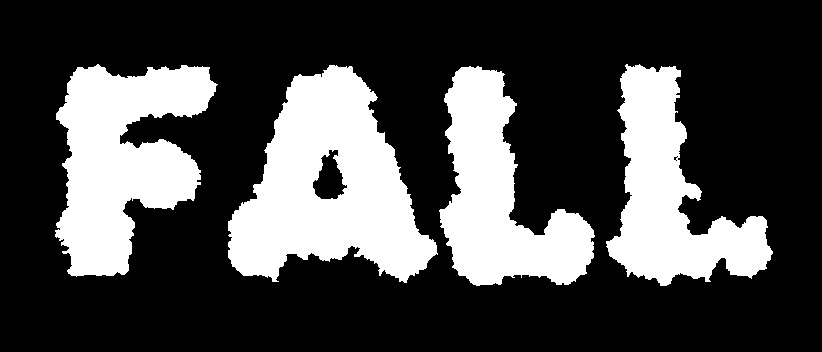}}
  \subfigure[\footnotesize{Source style $S'$}]{
  \includegraphics[width=0.48\linewidth]{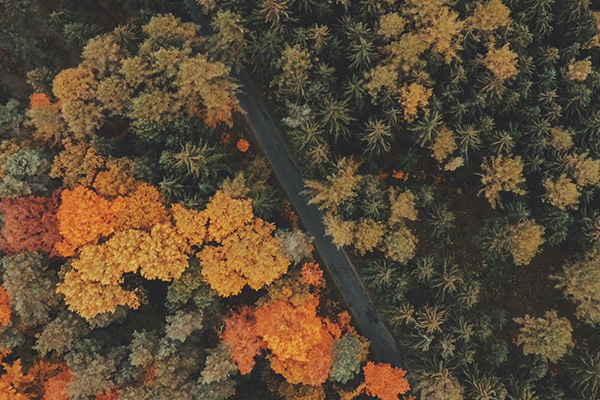}}
  \subfigure[\footnotesize{Our result}]{
  \includegraphics[width=0.48\linewidth]{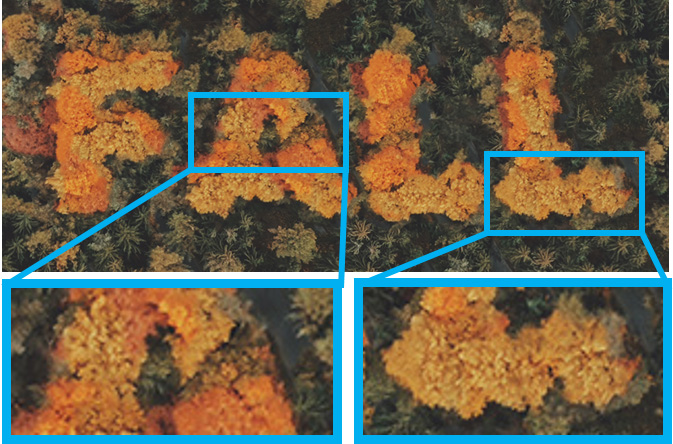}}
  \subfigure[\footnotesize{Result of Image Quilting~\cite{Efros2001Image}}]{
  \includegraphics[width=0.48\linewidth]{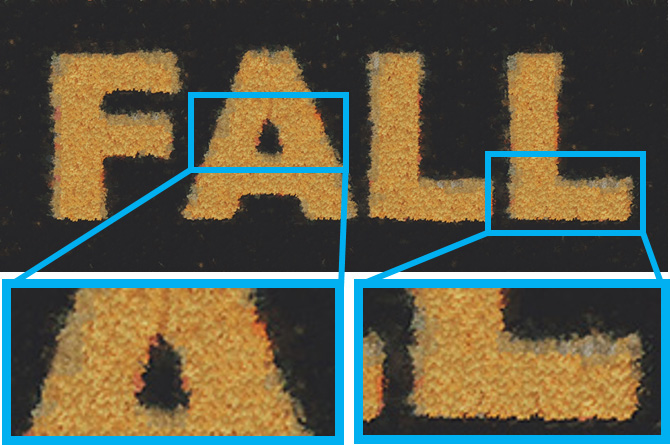}}
  \subfigure[\footnotesize{Result of Neural Style~\cite{gatys2016image}}]{
  \includegraphics[width=0.48\linewidth]{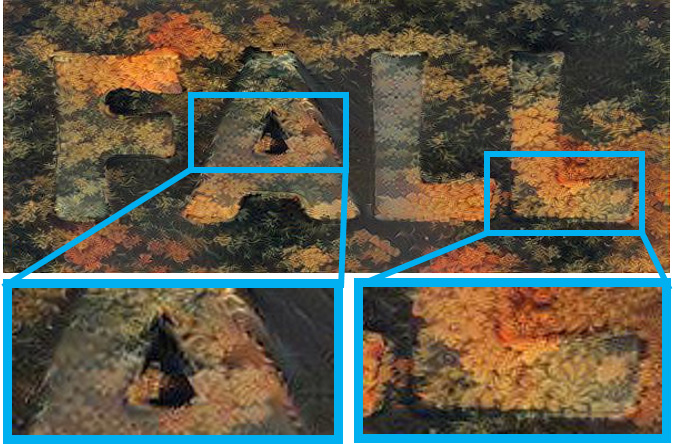}}
  \caption{Visual comparison with unsupervised style transfer methods. By structure transfer, our result better characterizes the shape of the forest canopy. Cropped regions are zoomed for better comparison.
  \textit{Image credits: Unsplash users Jakub Sejkora.}}\label{fig:experiment_shape_transfer}
\end{figure}

\subsection{Comparison of Style Transfer Methods}
\label{sec:comparison}

In Fig.~\ref{fig:experiment1}, we present a comparison of our method with six state-of-the-art supervised and unsupervised style transfer techniques on text stylization. For supervised methods, the structure guidance map $S$ extracted by our method in Section \ref{sec:foreground} is directly given as input. Please enlarge and view these figures on the screen for better comparison.

\textbf{Structural consistency}. In comparison to these approaches, our method can preserve the critical structural characteristics of textures in the source style image. Other methods do not consider to adapt the stroke contour to the source textures. As a result, they fail to guarantee structural consistency. For example, text boundaries in most methods are rigid in the \textit{leaf} group of Fig.~\ref{fig:experiment1}. By comparison, Neural Style~\cite{gatys2016image} and CNNMRF~\cite{Li2016Combining} implicitly characterize the texture shapes using deep-based features, while our method explicitly transfers structural features. Therefore only these three approaches create leaf-like letters. Similar cases can also be found in the \textit{spume} and \textit{coral reef} groups of Fig.~\ref{fig:experiment1}. The structural consistency achieved by our method can be better observed in the zoomed regions in Fig.~\ref{fig:experiment_shape_transfer}, where even Neural Style~\cite{gatys2016image} does not appear to transfer structure effectively.

\begin{figure}[t]
  \centering
  \subfigure[\footnotesize{Target text $T$}]{
  \includegraphics[width=0.48\linewidth]{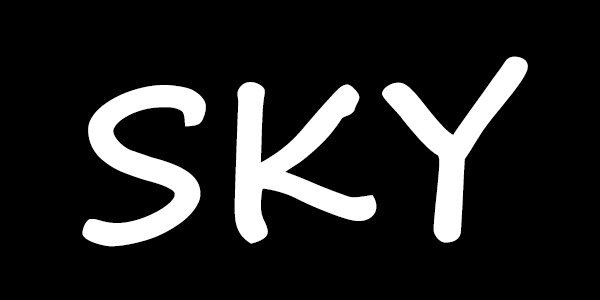}}
  \subfigure[\footnotesize{Result of Image Analogies~\cite{Hertzmann2001Image}}]{
  \includegraphics[width=0.48\linewidth]{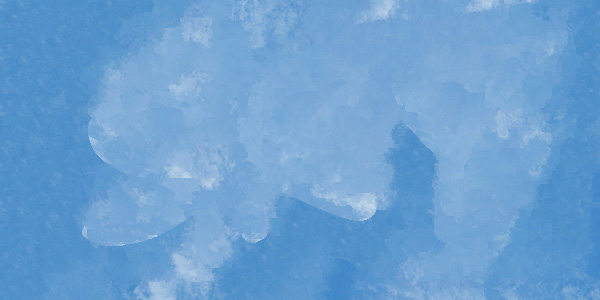}}\vspace{-1mm}
  \subfigure[\footnotesize{Result of Text Effects Transfer~\cite{Yang2017Awesome}}]{
  \includegraphics[width=0.48\linewidth]{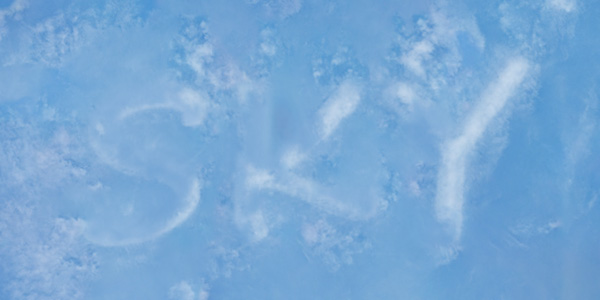}}
  \subfigure[\footnotesize{Our result}]{
  \includegraphics[width=0.48\linewidth]{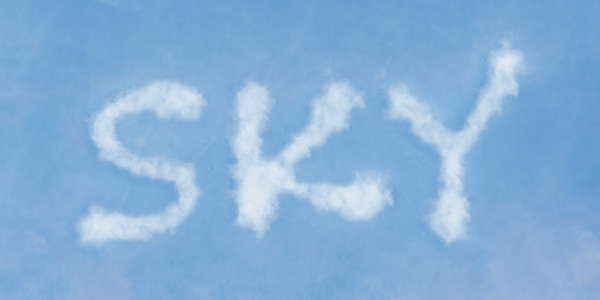}}
  \caption{Visual comparison with supervised style transfer methods. Our method yields the most distinct foreground against the background, thus well preserving shape legibility. In this example, $S'$ in Fig.~\ref{fig:saliency_transfer} is used.}\label{fig:experiment_saliency}
\end{figure}

\begin{figure}
  \centering
  \subfigure{
  \includegraphics[width=0.99\linewidth]{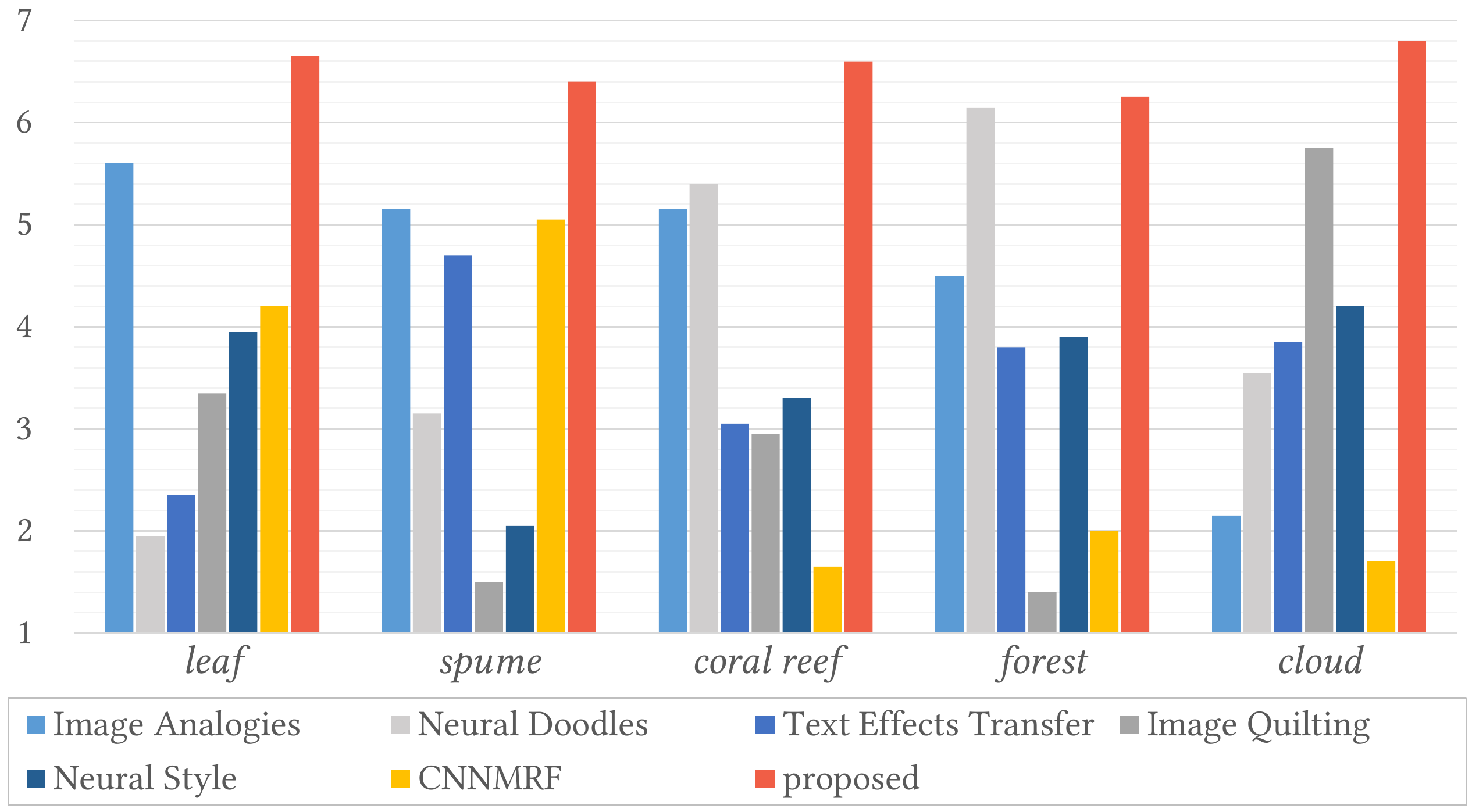}}
  \caption{Average evaluation scores of different methods for four test cases in Figures~\ref{fig:experiment1}-\ref{fig:experiment_saliency} from our 20-person study.}\label{fig:user-study1}
\end{figure}

\textbf{Text legibility}. For supervised style transfer approaches, the binary guidance map can only provide rough background/foreground constraints for texture synthesis. Consequently, the background of the \textit{island} result by Image Analogies~\cite{Hertzmann2001Image} finds many salient repetitive textures to fill, which confuses itself with the foreground.
Text Effects Transfer~\cite{Yang2017Awesome} introduces an additional distribution constraint for text legibility, which, however, is not effective for pure texture images. For example, due to the scale discrepancy between $S'$ and $T$, the distribution constraint forces Text Effects Transfer~\cite{Yang2017Awesome} to place textures compactly inside the text, leading to textureless artifacts in \textit{leaf} and \textit{spume} results.
Our method further proposes a complimentary saliency constraint, resulting in the creation of the artistic text that highlights from clean backgrounds. We show in Fig.~\ref{fig:experiment_saliency}
that when the foreground and background colors are not contrasting enough, our approach demonstrates greater superiority.

\begin{figure*}[t]
  \centering
  \subfigure{
  \includegraphics[width=0.99\linewidth]{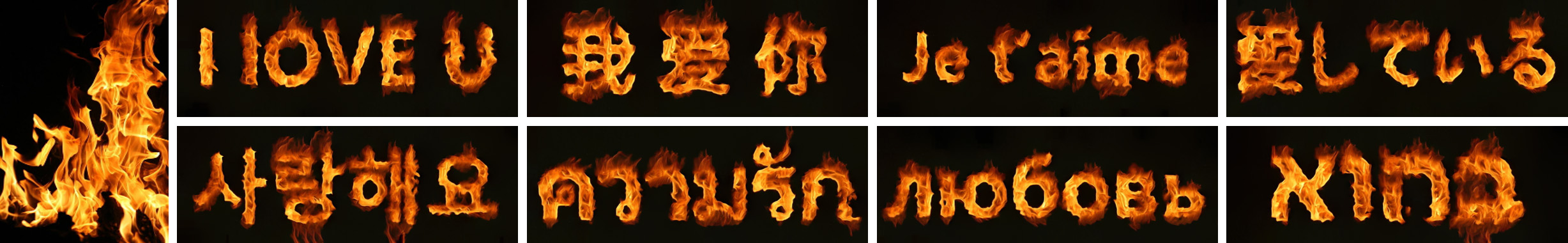}}
  \caption{Performance on different languages. The leftmost image is the source style $S'$.}\label{fig:language}
\end{figure*}
\begin{figure*}
  \centering
  \subfigure{
  \includegraphics[width=0.99\linewidth]{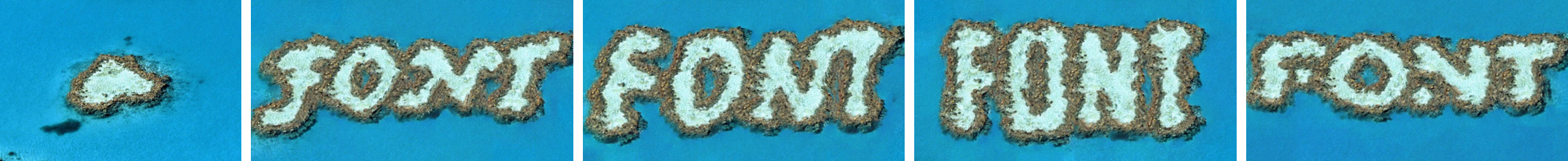}}
  \caption{Performance on different fonts. The leftmost image is the source style $S'$. \footnotesize\textit{Image credits: Unsplash users Yanguang Lan.}\normalsize}\label{fig:font}
\end{figure*}
\begin{figure*}
  \centering
  \subfigure[\footnotesize{\textit{barrier reef}}]{
  \includegraphics[height=0.27\linewidth]{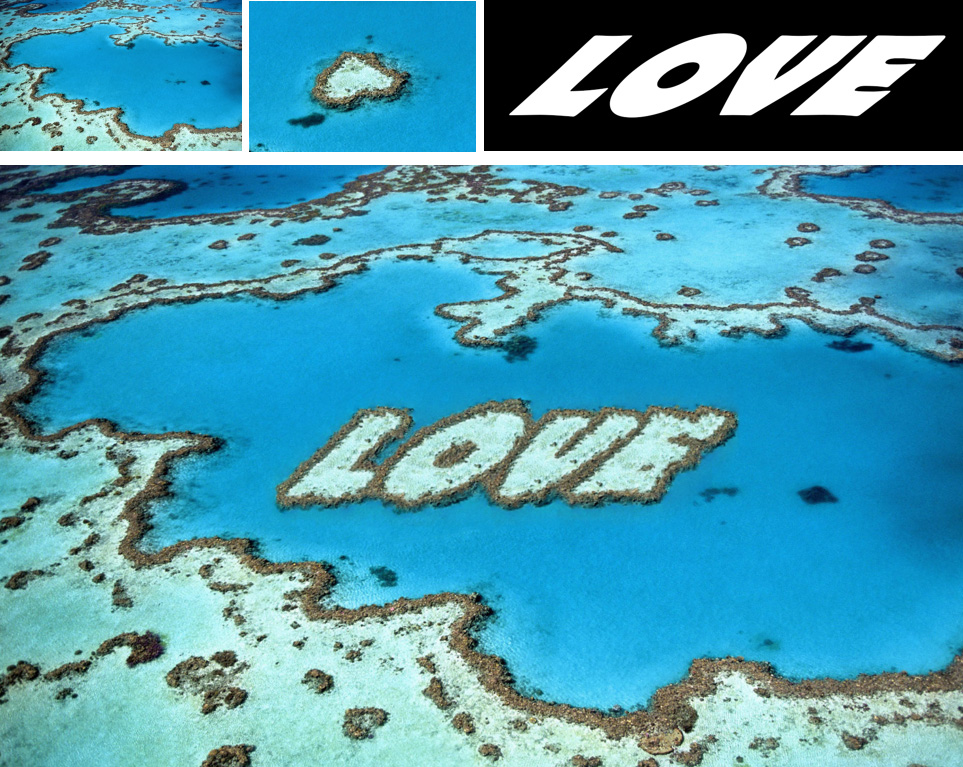}}
  \subfigure[\footnotesize{\textit{cloud}}]{
  \includegraphics[height=0.27\linewidth]{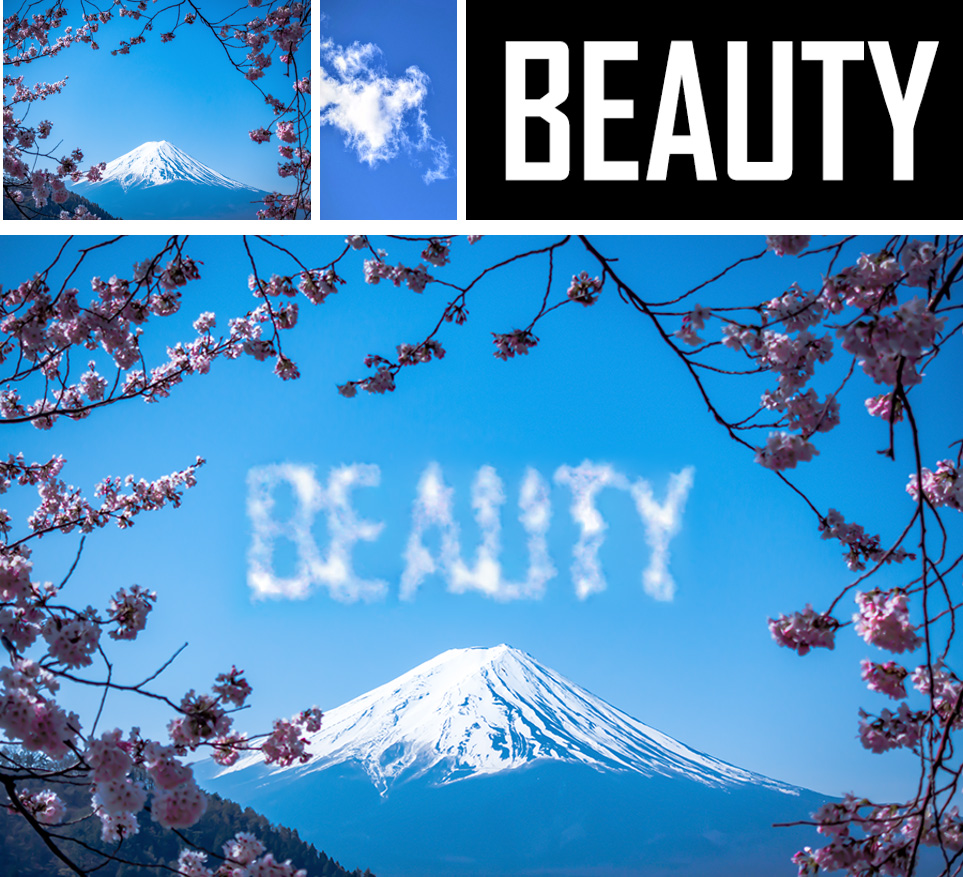}}
  \subfigure[\footnotesize{\textit{flame}}]{
  \includegraphics[height=0.27\linewidth]{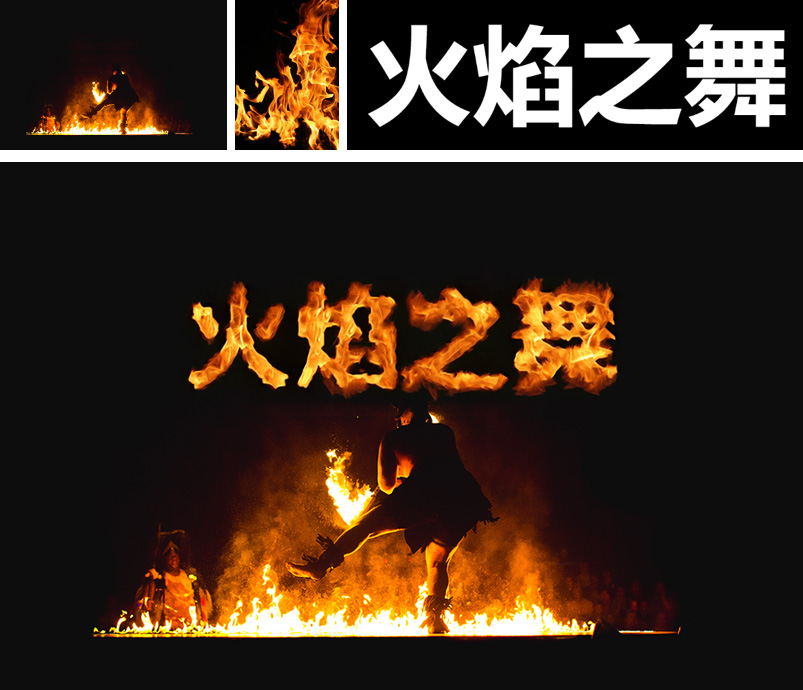}}\vspace{-1mm}
  \caption{Visual-textual presentation synthesis. For each result group, three images in the upper row are $I$, $S'$ and $T$, respectively. The lower one is our result. More examples can be found in the supplementary material.
  \textit{Image credits: Unsplash users Yanguang Lan, JJ Ying, Tim Gouw and Thomas Kelley.}}\label{fig:experiment3}
\end{figure*}

\begin{figure}
  \centering
  \subfigure{
  \includegraphics[width=0.99\linewidth]{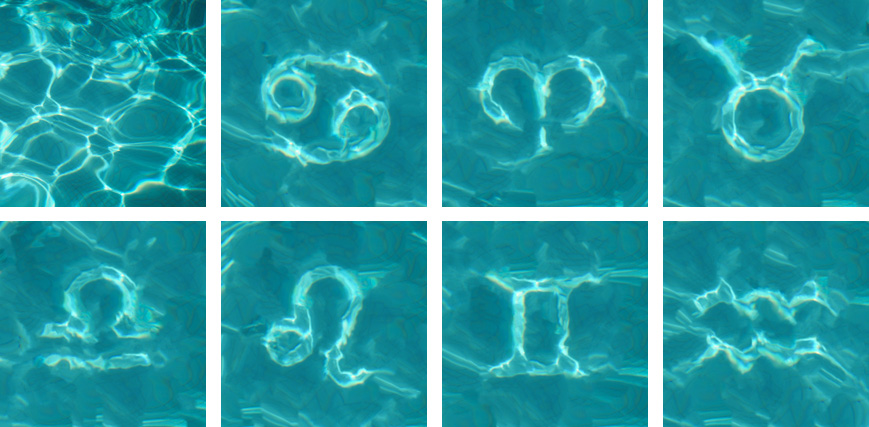}}
  \caption{Rendering rippling Zodiac symbols using a photo of water on the top left.
  \textit{Image credits: Unsplash users Rapha\"{e}l Biscaldi.}}\label{fig:constellation}
\end{figure}

\begin{figure}
  \centering
  \subfigure{
  \includegraphics[width=0.99\linewidth]{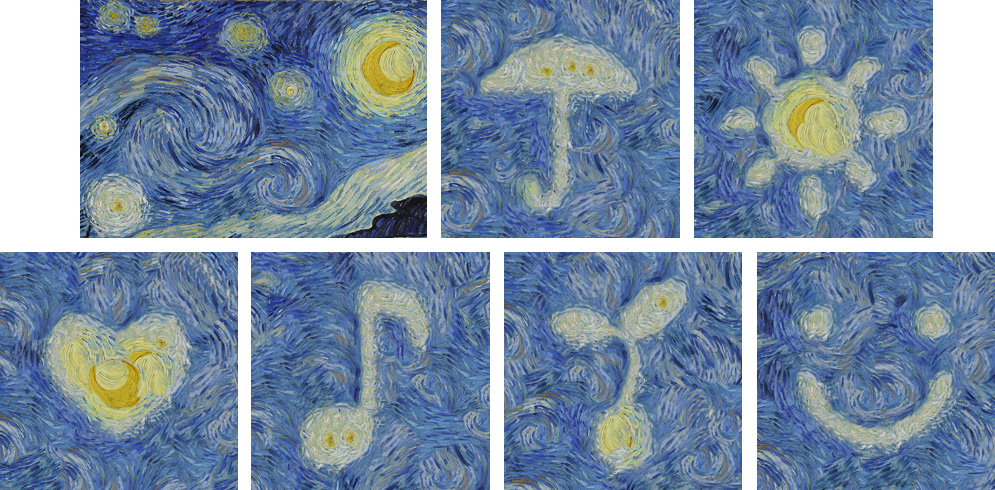}}
  \caption{Rendering emoji icons with the painting style of Van Gogh using ``\textit{The Starry Night}'' on the top left.}\label{fig:symbol}
\end{figure}

\textbf{Texture naturalness}. Compared with other style transfer methods, our method produces visually more natural results. For instance, our method places irregular coral reefs of different densities based on the shape of the text in the example \textit{coral reef} of Fig.~\ref{fig:experiment1}, which highly respects the contents of $S'$. This is achieved by our context-aware style transfer to ensure structure, color, texture and semantic consistency. By contrast, Image Quilting~\cite{Efros2001Image} relies on patch matching between two completely different modalities $S'$ and $T$, thus its results are just as flat as the raw text. Three deep-learning based methods, Neural Style~\cite{gatys2016image}, CNNMRF~\cite{Li2016Combining} and its supervised version Neural Doodles~\cite{Champandard2016Semantic}, transfer suitable textures onto the text. However, their main drawbacks are the color deviation and checkerboard artifacts (see example \textit{coral reef} in Fig.~\ref{fig:experiment1}).

\textbf{User study}. For quantitative evaluation, we conducted user studies where twenty participants were shown five test cases in Figures~\ref{fig:experiment1}-\ref{fig:experiment_saliency} and were asked to assign 1 to 7 scores to the seven methods in each case (a higher score indicates that the stylized result is more consistent in style with the style image). Figure~\ref{fig:user-study1} shows that our method outperforms other methods in all cases, obtaining the best average score of 6.54, significantly higher than 4.51, 4.04, 3.55, 2.99, 3.48 and 2.92 of Image Analogies~\cite{Hertzmann2001Image}, Neural Doodles~\cite{Champandard2016Semantic}, Text Effects Transfer~\cite{Yang2017Awesome}, Image Quilting~\cite{Efros2001Image},  Neural Style~\cite{gatys2016image} and CNNMRF~\cite{Li2016Combining}, respectively.

\subsection{Generating Stylish Text in Different Fonts and Languages}
\label{sec:more_result}

We experimented on text in various fonts and languages to test the robustness of our method. Some results are shown in Figs.~\ref{fig:language}-\ref{fig:font}. The complete results can be found in the supplementary material.
In Fig.~\ref{fig:language}, characters are quite varied in different languages. Our method successfully synthesizes dancing flames onto a variety of languages, while maintaining their local fine details, such as the small circles in Thai.
In Fig.~\ref{fig:font}, the rigid outlines of the text are adjusted to the shape of a coral reef, without losing the main features of its original font. Thanks to our stroke trunk protection mechanism, our approach balances the authenticity of textures with the legibility of fonts.

\subsection{Visual-Textual Presentation Synthesis}
\label{sec:typography_design}

\begin{figure}[t]
  \centering
  \subfigure{
  \includegraphics[width=0.99\linewidth]{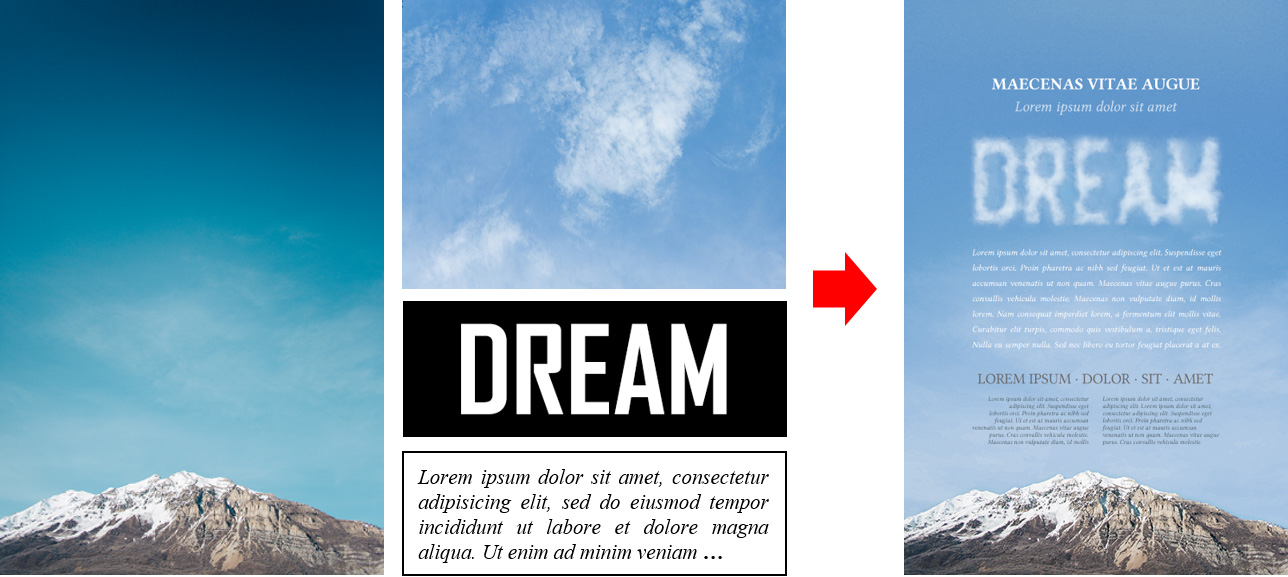}}
  \caption{Application: Computer aided poster design.
  \textit{Image credits: Unsplash users Evan Kirby, Ashim D'Silva.}}\label{fig:poster}
\end{figure}

We aim to synthesize professional looking visual-textual presentation that combines beautiful images and overlaid stylish
text. In Fig.~\ref{fig:experiment3}, three visual-textual presentations automatically generated by our method are
provided. In the example \textit{barrier reef}, a LOVE-shaped barrier reef is created, which is visually consistent with the background photo. We further show in the example \textit{cloud} that we can integrate completely new elements into the background. Clouds with a specific text shape are synthesized in the clear sky. The colors in the sky of $S'$ are adjusted to match those in the background, which effectively avoids abrupt image boundaries. Please note the text layout automatically determined by our method is quite reasonable. Therefore, our approach is capable of artistically embellishing photos with meaningful and expressive text and symbols, thus providing a flexible and effective tool to create original and unique visual-textual presentations. This art form can be employed in posters, magazine covers and many other media. We show in Fig.~\ref{fig:poster} a poster design example, where its stylish headline is automatically generated by our method and the main body is manually designed. A headline made of clouds effectively enhances the attractiveness of the poster.

\subsection{Symbol and Icon Rendering}
\label{sec:application}

The proposed method has the ability to render textures for text-based geometric shapes such as symbols and icons.
Fig.~\ref{fig:constellation} shows that our method successfully transfers rippling textures onto the binary Zodiac symbols.
It seems that the proposed method is also capable of stylizing more general shapes, like the emoji icons in Fig.~\ref{fig:symbol}.
Meanwhile, we notice that our saliency term selects the prominent orange moon to be synthesized into the \textit{sun} and \textit{heart}, which enriches the color layering of the results.

\begin{figure}[t]
  \centering
  \subfigure[input]{
  \includegraphics[width=0.31\linewidth]{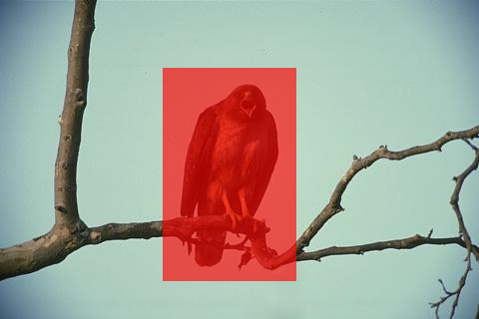}}
  \subfigure[inpainting result]{
  \includegraphics[width=0.31\linewidth]{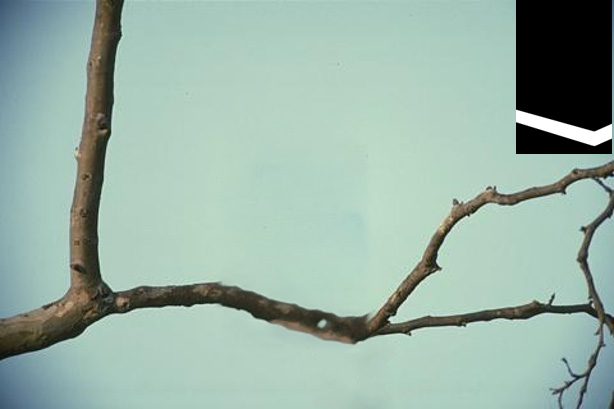}}
  \subfigure[inpainting result]{
  \includegraphics[width=0.31\linewidth]{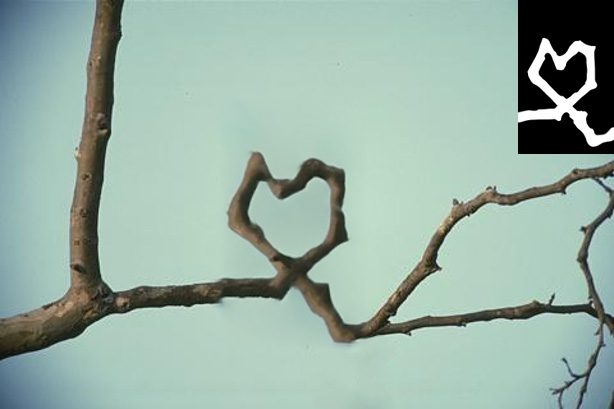}}
  \caption{Structure-guided image inpainting.}\label{fig:inpainting}
\end{figure}

\subsection{Structure-Guided Image Inpainting}
\label{sec:application}

Our method can be naturally extended to structure-guided image inpainting. Fig.~\ref{fig:inpainting} demonstrates the feasibility of controlling the inpainting result via user-specified shapes. The input photo in Fig.~\ref{fig:inpainting}(a) is used as both a style image and a background image, where its red mask directly indicates the embedding position.
The user sketches in white the approximate shapes of the branches (shown in the upper right corner of the result), and the resulting sketch serves as target $T$ for stylization. Figs.~\ref{fig:inpainting}(b)(c) are our inpainting results, which are quite plausible and the filled regions blend well with the background.

\newcommand{\tabincell}[2]{\begin{tabular}{@{}#1@{}}#2\end{tabular}}

\begin{table*} [t]
\caption{Running time (seconds) of the proposed method}
\label{tb:running_time}
\centering
\begin{tabular}{l|ccc|ccccc|c}
\hline
Test set  & $T$ & $S'$ & $I$ & \tabincell{c}{Guidance\\extraction}  & \tabincell{c}{Position\\estimation} & \tabincell{c}{Color\\transfer} & \tabincell{c}{Structure\\transfer} & \tabincell{c}{Texture\\transfer} & Total \\
\hline
\textit{barrier reef} & 975$\times$308 & 566$\times$377 & 1920$\times$1200 & 11.08 & 28.11 & - & 6.43 & 255.93 & 301.55 \\
\textit{cloud} & 477$\times$197 & 500$\times$750 & 900$\times$600 & 16.86 & 21.76 & 15.23 & 6.44 & 247.37 & 307.85 \\
\textit{flame} & 916$\times$300 & 400$\times$575 & 1556$\times$1024 & 15.87 & 13.63 & - & 17.37 & 314.15 & 361.02 \\
\hline
Averaged & 0.22Mp & 0.27Mp & 1.5Mp & 14.60 & 21.17 & 15.23 & 10.15 & 272.48 & 323.47 \\
\hline
\end{tabular}
\end{table*}

\subsection{Running Time}
\label{sec:runningtime}

When analyzing the complexity of the proposed method, we consider the time of guidance map extraction, position estimation and color/structure/texture transfer. To simplify the analysis, we assume the target image $T$ has $N$ pixels, and the image resolution of $S'$ and $I$ have the same magnitude $O(N)$ as $T$. In addition, the patch size and the number of iterations are constants that can be ignored in computational complexity.

\textit{Guidance map extraction}. According to~\cite{xu2012structure,achanta2012slic,zhang2013saliency}, the complexity of RTV, super pixel extraction and saliency detection is $O(N)$. K-means has a practical complexity of $O(NKt)$~\cite{elkan2003using}, where $K=2$ and $t$ is the number of iterations. Ignoring $K$ and $t$, the total complexity of guidance map extraction is $O(N)$.

\textit{Position estimation}. The complexity of calculating $U_v$, $U_s$, $U_a$ and box filter is $O(N)$. For coherence cost $U_c$, we use FLANN~\cite{muja2014scalable} to search matched patches between $S'$ and $I$, which has a complexity of $O(N\log N)$. Therefore, the overall complexity of the proposed position estimation is $O(N\log N)$.

\textit{Style transfer}. According to~\cite{hertzmann2001algorithms}, color transfer has an $O(N)$ complexity. During structure transfer, patches along the shape boundary are matched using FLANN~\cite{muja2014scalable}. The upper bound of the patch number is $O(N)$ and thus the proposed structure transfer is $O(N\log N)$ complex. As reported in~\cite{PatchMatch}, PatchMatch in texture transfer has a complexity of $O(N\log N)$.

In summary, the overall computational complexity of the proposed method is $O(N\log N)$.

Table~\ref{tb:running_time} shows the running time of our method on three test sets (Fig.~\ref{fig:experiment3}) with Intel Xeon 3.00 GHz CPU E5-1607. The proposed method is implemented on MATLAB platform. Texture transfer is our major computational bottleneck, which accounts for about $85\%$ of the total time. This is because matching patches in mega-pixel (Mp) images can be slow at finer scales. As our method is not multithreaded, it just uses a single core. Our method can be further speeded up by implementing a well-tuned and fully parallelized PatchMatch algorithm.

\begin{figure}[t]
  \centering
    \includegraphics[width=0.95\linewidth]{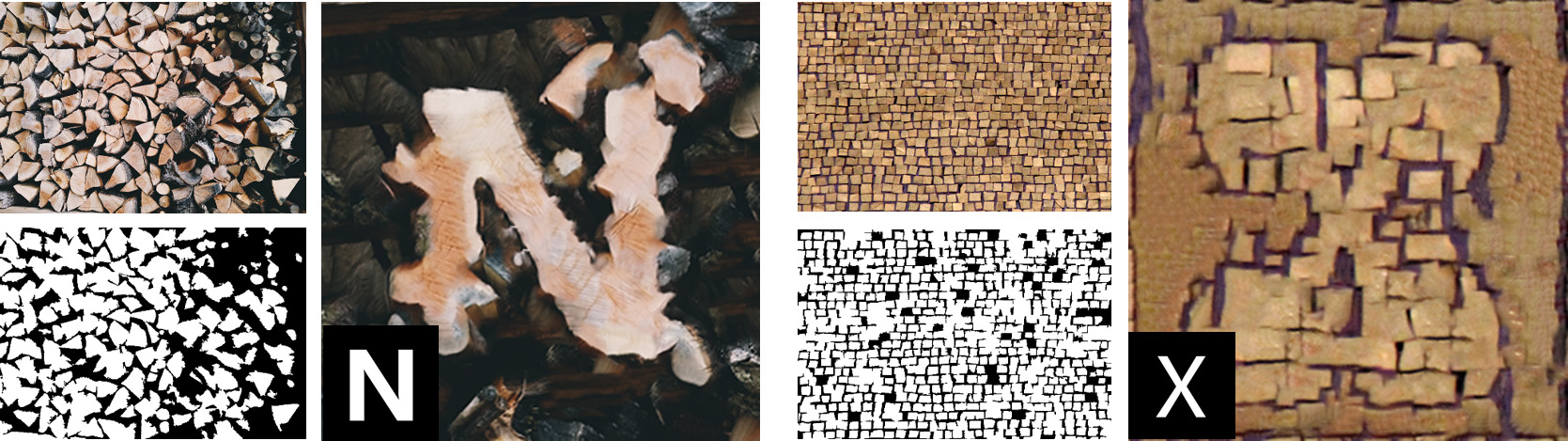}
  \caption{Some examples of failure cases.
  \textit{Image credits: Unsplash users Brigitte Tohm, Edouard Dognin.}}\label{fig:limitation}
\end{figure}

\subsection{Limitation}
\label{sec:limitation}

While our approach has generated visually appealing results, some limitations still exist.
Our guidance map extraction relies on simple color features, and is not fool-proof.
Main structure abstraction from complex scenarios might be beyond its capability. The problem may be addressed by either employing high-level deep features or user interactions. Moreover, even with precise guidance maps, our method may fail when the source style contains tightly spaced objects. As shown in Fig.~\ref{fig:limitation}, our method yields interesting results, but they do not correctly reflect the original texture shapes.
The main reason is that it is hard for our patch-based method to find pure foreground or background patches in the dense patterns for shape and texture synthesis. Therefore, when synthesizing the foreground (background) region, shapes/textures in the background (foreground) region will be used, which causes mixture and disorder.

\section{Conclusion and Future Work}
\label{sec:conclusion}

In this paper, we demonstrate a new technique for text-based binary image stylization and synthesis to incorporate binary shape and colorful images. We exploit guidance map extraction to facilitate the structure and texture transfer. Cues for seamlessness and aesthetics are leveraged to determine the image layout. Our context-aware text-based image stylization and synthesis approach breaks through a barrier between images and shapes, allowing users to create fine artistic  shapes and to design professional looking visual-textual presentations.

There are still some interesting issues for further investigation. A direction for future work is the automatic style image selection. $S'$ that shares visual consistency and semantic relevance with the background image will contribute more to seamless embedding and aesthetic interests. Recommendation of $S'$ could be achieved by leveraging deep neural networks to extract semantic information.




\bibliographystyle{IEEEtran}
\bibliography{bibliography}

\ifCLASSOPTIONcaptionsoff
  \newpage
\fi

\end{document}